# Tightly Joined Positioning and Control Model for Unmanned Aerial Vehicles Based on Factor Graph Optimization

Peiwen Yang, Weisong Wen*, *Member, IEEE*, Shiyu Bai, *Member, IEEE*, and Li-Ta Hsu, *Senior Member, IEEE*

***Abstract*—The existing motion control pipeline, where the positioning and control are decoupled, struggles to adapt to epistemic and aleatoric uncertainty and nonlinear dynamics. As a result, the motion control reliability of the unmanned aerial vehicle (UAV) is significantly challenged in complex dynamic areas. For example, the ubiquitous global navigation satellite system (GNSS) positioning can be degraded by the signal reflections from surrounding high-rising buildings in complex urban areas, leading to significantly increased positioning uncertainty. Given that positioning and control are highly correlated, this research proposes a tightly joining positioning and control model (JPCM) based on factor graph optimization (FGO). Specifically, the sensor measurements are formulated as the factors in the probabilistic factor graph. In addition, the model predictive control (MPC) is also formulated as the additional factor in the probabilistic factor graph. The factor graph contributed by both the positioning-related factors and the MPC-based factors deeply exploits the complementariness of positioning and control. Finally, we validate the effectiveness and resilience of the proposed method using simulations and real-world experiments that show significantly improved trajectory following performance. To benefit the research community, we open-source our code and make it available at https://github.com/RoboticsPolyu/IPN_MPC.**

LIST OF SYMBOLS

| Symbol | Remarks |
|---|---|
| $\boldsymbol{\omega}^{\times}$ ($\boldsymbol{\omega} = [\omega_1, \omega_2, \omega_3]^{\mathrm{T}}$) | $\boldsymbol{\omega}^{\times} = \begin{bmatrix} 0 & -\omega_3 & \omega_2 \\ \omega_3 & 0 & -\omega_1 \\ -\omega_2 & \omega_1 & 0 \end{bmatrix}$ |
| $\mathbf{x} = [\mathbf{p}^{\mathrm{T}}, \boldsymbol{\theta}^{\mathrm{T}}, \mathbf{v}^{\mathrm{T}}, \boldsymbol{\omega}^{\mathrm{T}}]^{\mathrm{T}}$ | State $\mathbf{x}$, position $\mathbf{p}$, rotation $\mathbf{R} = \exp(\boldsymbol{\theta}^{\times})$, velocity $\mathbf{v}$, and angular speed $\boldsymbol{\omega}$ |
| $\hat{\mathbf{x}}$ | Optimized state |
| $\Delta\mathbf{x}$ | State increment w.r.t initial state |
| $\delta\mathbf{x}$ | State increment in the tangent space |
| $\mathbf{z}$ | Sensor measurement |
| $\mathbf{u}$ | Control input |
| $\mathbf{z}^{\mathrm{r}}, \mathbf{u}^{\mathrm{r}}$ | Reference state and reference input |
| $\mathbf{T}, \mathbf{M}$ | Resultant thrust $\mathbf{T}$ and resultant torque $\mathbf{M}$ |
| $\mathbf{P}$ | Covariance matrix |
| $\mathbf{D}$ | Aerodynamic drag coefficient matrix |
| $\mathbf{I}_b$ | Moment of inertia |
| $\mathrm{g}$ | Gravity constant |
| $\mathrm{m}_b$ | UAV's mass |
| $\mathrm{c}_t$ | Rotor's aerodynamic thrust coefficient |
| $\mathrm{k}_m$ | Rotor's aerodynamic torque coefficient |
| $\mathbf{p}_{r_j}^{b}$ | $j$-th rotor's position in the body frame |
| $\mathbf{T}_{b_l}^{w}$ | UAV's pose in the world frame at timestamp $l$ |
| $\mathbf{T}_{l+1}^{l}$, $\mathbf{P}_{l}^{L}$ | Relative pose transformation from timestamp $l$ to timestamp $l+1$ and its covariance matrix |
| $\mathbf{r}^{R}$,$\mathbf{r}^{A}$,$\mathbf{r}^{ref}$, $\mathbf{r}^{D}$,$\mathbf{h}(\mathbf{u}_j)$,$\mathbf{r}^{U}$ | Relative factor residual, absolute factor residual, reference trajectory factor residual, dynamic control factor residual, control bound factor residual, and control rate factor residual |
| $\mathbf{Q}_k$, $\mathbf{Q}_N$, $\mathbf{R}_t$, $\mathbf{Q}_{\mathrm{lim}}$ | Weighting matrix in MPC; covariance matrix in JPCM |
| $\mathbf{I}_n$ | n-dimensional unit vector |

The authors are with the Department of Aeronautical and Aviation Engineering, Polytechnic University, Hong Kong, E-mail: (peiwen1.yang@connect.polyu.hk; welson.wen@polyu.edu.hk; shiyu.bai@polyu.edu.hk; lt.hsu@polyu.edu.hk). (*Corresponding author: Weisong Wen.*)

## I. Introduction

<u>*The pipeline of decoupling positioning and control has dominated the practice of UAVs for decades*</u>: Reliable motion control is of great significance for the applications of unmanned aerial vehicles (UAV) [1]. The positioning can be achieved using multi-sensory fusion [2, 3]. Then the state estimation from the positioning module is fed to the controller of the UAV, such as the popular model predictive control (MPC) [4-6]. However, existing control models are limited in terms of reasoning about sources of epistemic uncertainty (e.g. uncertain disturbances) and aleatoric uncertainty (e.g. measurement's randomness) [7], and nonlinear dynamics. Therefore, we encode these uncertain sources as probabilistic factor graphs [8].

<u>*Positioning solutions can be challenged with large uncertainty in complex scenes*</u>: In particular, the positioning can be derived from the state estimator, such as the Kalman filter [9, 10] or factor graph optimization (FGO) [11, 12] based on the onboard sensors. For example, the global navigation satellite system (GNSS) was widely used in vehicular localization under open sky and usually combined with proprioceptive sensors such as inertial measurement unit (IMU) and exteroceptive sensors such as light detection and ranging (LiDAR) during satellite signal outages [3, 13-18]. However, the performance of these solutions relies strongly on the environmental conditions [7, 19, 20]. On the one hand, the GNSS positioning can be significantly degraded in urban canyons due to the multipath effects [20]. Although numerous works were proposed to mitigate the impacts of the GNSS outliers [20, 21], the positioning solution derived from the GNSS is still unreliable in highly urbanized areas with significant uncertainty. Moreover, the LiDAR-based positioning can also be significantly challenged in urban canyons due to the excessive dynamic objects [22], introducing uncertainty to the control system.

*Insufficient awareness of the positioning uncertainty can challenge the controllers in complex scenes*: Usually, the positioning solution is directly fed to the controller, such as the PID controller [23], assuming a perfect positioning accuracy is guaranteed. Unfortunately, positioning accuracy is usually not guaranteed in complex scenes using the existing GNSS or multi-sensory integration [3, 13-18]. As a result, the controller would inevitably lead to non-optimal control commands, causing dangerous UAV actions. Hence, control stability under large positioning uncertainty is essential for safety. Estimating the uncertainty of a state in a controller typically involves using state observers or estimators [24]. In traditional control frameworks, positioning, observer, and control are decoupled and cannot simultaneously adapt to the uncertainty of state [24] and nonlinear dynamics in a single problem [25]. Instead of carefully coping with the uncertainty arising from the positioning module, the existing work [5, 26] mainly focuses on the disturbances of the dynamical model, while ignoring the uncertainties of raw measurements. However, it is difficult to reliably distinguish the positioning uncertainty and the exact disturbance coming from the surrounding environments. To effectively consider the uncertainty of the positioning module, researchers proposed to directly add the positioning uncertainty as the covariance matrix of the cost function in the MPC [27, 28], which has attracted much attention recently. However, the work in [27, 28] has several key drawbacks: (1) it relied heavily on the reliability of the uncertainty provided by the positioning module, which is hard to obtain in complex scenes as mentioned above. (2) the information from the controller cannot flow back to the positioning modules. For example, the dynamics of the UAV system applied in the controller can also contribute to the positioning modules [29, 30]. In other words, the positioning modules and the controller are highly complementary and correlated.

*Therefore, we investigate the integration of* positioning and control. As illustrated in Fig 1, the classical motion control framework's pipeline runs in cycles, first estimating the state by positioning and then deriving a control input by MPC. Interestingly, the work from Prof. Frank Dellaert, from the School of Interactive Computing at the Georgia Institute of Technology, U.S., proposed a positioning and planning joint model [31] for robotics, in a tightly coupled manner. As a result, the uncertainty from both the positioning and the path planning can be considered optimally and simultaneously. The work [32, 33] reformulated the measurements over a fixed time window into an optimal control problem instead of a single measurement for the estimation at each time step. Similar work was done in equality and inequality-constrained optimal control [34, 35], and kinematic and dynamics problems for multi-body systems [36], which were solved by factor graphs [6]. Besides, some works [37, 38] tightly integrated trajectory estimation and path planning by utilizing obstacle avoidance factors and goal factors. The above papers revealed the feasibility of applying factor graphs to trajectory planning or control. *All these works inspire us to propose a tightly coupled model to solve the positioning and control problems simultaneously.*

*Contribution:* In this paper, we proposed a tightly joined positioning and control model (JPCM) that integrates positioning and control using a unified quadratic problem. First, positioning and control problems are formulated as a combined problem using a unified factor graph, ensuring efficient information flow between each other. Then, the positioning cost function from sensor-based measurements and control cost function from MPC-based constraints are converted into a tightly joined factor graph. Then, the joined FGO problem is solved by the popular GTSAM [11] library. Finally, the JPCM is evaluated by a well-designed quadrotor simulator, which considers the impact of aerodynamic drag force and the actuator's dynamic [39]. Then, lemniscate trajectory tracking experiments are conducted to demonstrate the proposed JPCM.

Specifically, the main contributions of this paper include:

(1) *Propose a tightly coupled model to solve motion control problems, combining positioning and control constraints into a probability factor graph.* We categorize measurements into absolute factors, such as the GNSS factor, and relative factors, like the LiDAR scan-matching factor. Additionally, we formulate dynamics, trajectory following, and control input constraints as control-based factors. All these factors are consolidated into a unified factor graph, which is then solved to yield the state and the control input.

*(2)* *Eliminate the impact of aerodynamic drag force, JPCM-drag is proposed to tackle the path following offset.* The dynamics model of UAV in the proposed model considers aerodynamic drag.

(3) *Evaluate the proposed JPCM through trajectory tracking simulations and experiments.* Except for the uncertainty of sensors and dynamic processes, the sharp wind, and aerodynamic drag are also simulated to evaluate the robustness of the proposed JPCM. In addition, the feasibility of the proposed JPCM is verified using a quadrotor by real-time trajectory tracking experiments. Lastly, we have made the code open source to benefit the research community.

The structure of the paper is as follows. Section II introduces the UAV's dynamic model and uniformly formulates positioning and MPC using the same mathematical representation. Section III proposes the JPCM and its key components. In Section IV, numerical simulation and real-world experiments are conducted to evaluate the performance of the proposed model. Section V discusses more details. Finally, the conclusion is summarized in Section VI.

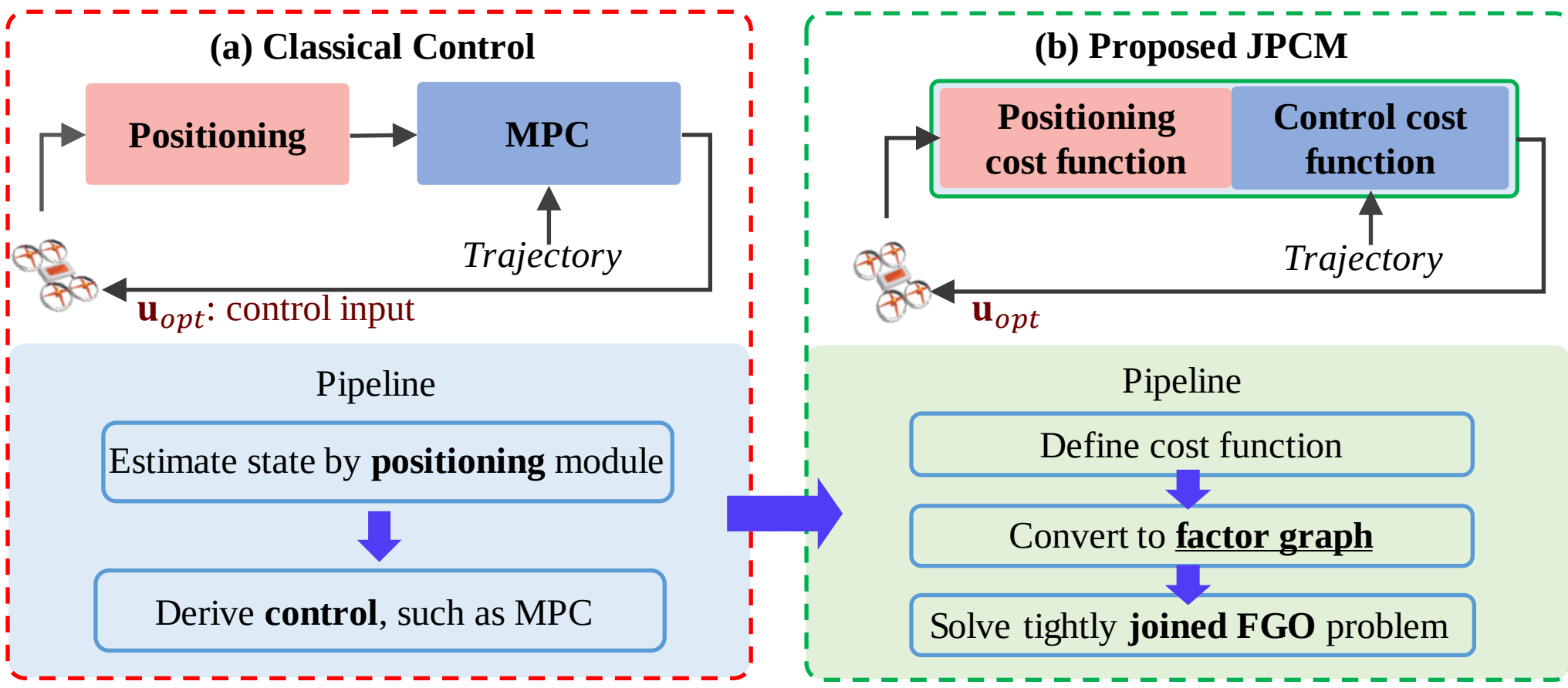

Fig. 1. The motion control model: from (left) the classical model-based control to (right) the proposed tight JPCM. (a) The model-based motion control method has a fixed loop. The positioning module and control module are decoupled. (b) The proposed JPCM integrates positioning and control into a cost function. Then, we convert the cost function to a probability factor graph, in which the positioning-related factors and MPC-based factors are tightly combined.

## II. Problem Description

As shown in Fig 2, both positioning (red box) and MPC (blue box) can be conceptualized as a factor graph on manifold [4, 8, 40]. In the context of a canonical positioning model, factors derived from measurements can be categorized into two distinct types [41]: absolute factors and relative factors. Absolute factors apply constraints to the robot's state at a specific moment, while relative factors constrain the variation between states at varying time intervals. For example, the measurement of IMU and LiDAR can construct relative factors. $\mathbf{x}_k$, $\hat{\mathbf{x}}_k$, and $\mathbf{z}^r$ is the initial state, estimated state, and reference state, respectively. In positioning problem, $\hat{\mathbf{x}}_k$ is constrained by sensors' measurements. $\delta\mathbf{x}$ is the state increment in the tangent space. In MPC, $\hat{\mathbf{x}}_k$ is constrained by the initial state $\mathbf{x}_0$, dynamics and reference state $\mathbf{z}^r$. In practical systems, the initial state $\mathbf{x}_0$ in MPC comes from the positioning solution, but the uncertainty of the positioning solution is ignored. As illustrated in Fig 2 (b), starting from an initial state $\mathbf{x}_0$, contains constraints $\mathbf{r}^{ref}$ from the reference states $\mathbf{z}^r$, constraints $\mathbf{r}^D$ from the dynamic model, and control input constraints. The dynamic factors $\mathbf{r}^D$ become the informational bridge. Mathematically, both positioning and control can be represented through factor graphs.

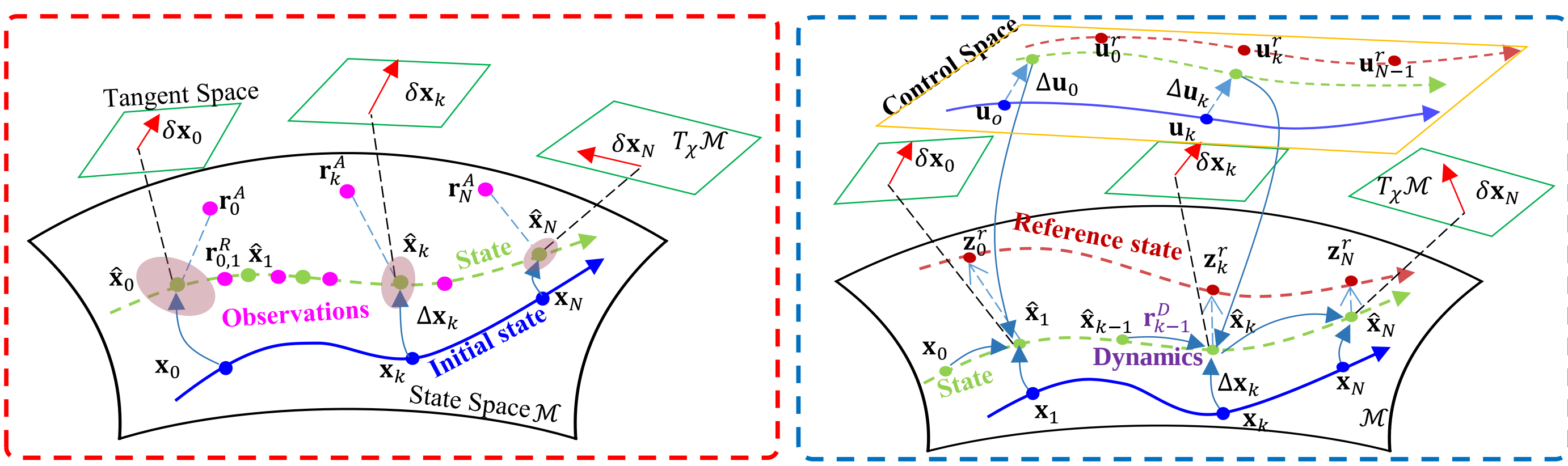

(a) Positioning

(b) MPC

Fig. 2. The representation and linearization of the positioning and MPC problem. (a) Positioning is a factor graph with absolute factors $\mathbf{r}^A$ and relative factors $\mathbf{r}^R$, which will be discussed in Section II-B. (b) MPC can also be formulated as a factor graph.

### *A. The UAV dynamics model*

The generalization dynamics model [24, 42] of a UAV is as follows:

$$\dot{\mathbf{x}} = \mathrm{f}(\mathbf{x}, \boldsymbol{\tau}) \tag{1a}$$

$$\boldsymbol{\tau} = \mathrm{g}(\mathbf{u}) \tag{1b}$$

where $\mathbf{x} = [(\mathbf{p}_b^w)^{\mathrm{T}}, (\boldsymbol{\theta}_b^w)^{\mathrm{T}}, (\mathbf{v}_b^w)^{\mathrm{T}}, (\boldsymbol{\omega}_b)^{\mathrm{T}}]^{\mathrm{T}}$ is the UAV's state that includes position $\mathbf{p}_b^w$, rotation $\mathbf{R}_b^w = \exp\{(\boldsymbol{\theta}_b^w)^{\times}\}$, velocity $\mathbf{v}_b^w$, and angular speed $\boldsymbol{\omega}_b$, and $\boldsymbol{\tau} = [\mathbf{T}_b^{\mathrm{T}}, \mathbf{M}_b^{\mathrm{T}}]^{\mathrm{T}}$ represents the resultant thrust $\mathbf{T}_b \in \mathbb{R}^3$ and resultant torque $\mathbf{M}_b \in \mathbb{R}^3$, and $\mathbf{u} = [u_0, \ldots, u_j, \ldots, u_K]^{\mathrm{T}}, K \geq 3, K \in \mathbb{N}$ is the control input vector, where $u_j$ represents the angular speed of the $j$-th motor of the UAV. $\mathrm{f}(*)$ and $\mathrm{g}(*)$ represent the generalization dynamics model and control allocation model, respectively.

The continuous-time linearized dynamics of (1a) is as follows [24, 42]:

$$\begin{aligned}\dot{\mathbf{p}}_b^w &= \mathbf{v}_b^w \\ \dot{\mathbf{R}}_b^w &= \mathbf{R}_b^w \boldsymbol{\omega}_b^\times \\ \dot{\mathbf{v}}_b^w &= -\mathbf{R}_G^w \mathbf{e}_3 \mathrm{g} + (\mathbf{R}_b^w \mathbf{T}_b + \mathbf{R}_b^w \mathbf{D} \mathbf{R}_w^b \mathbf{v}_b^w)/\mathrm{m}_b \\ \dot{\boldsymbol{\omega}}_b &= \mathbf{I}_b^{-1}(-\boldsymbol{\omega}_b^\times \mathbf{I}_b \boldsymbol{\omega}_b + \mathbf{M}_b)\end{aligned} \quad (2)$$

where $\mathrm{m}_b$ is UAV's mass, $\mathbf{e}_3 = [0,0,1]^{\mathrm{T}}$, g is the gravity constant, $\mathbf{D}$ is the aerodynamic drag coefficient matrix [39], and $\boldsymbol{\omega}_b^\times$ represents an operator that spins a 3D vector $\boldsymbol{\omega}_b$ to a skew-symmetric as follows:

$$\boldsymbol{\omega}_b^\times = \begin{bmatrix} 0 & -\omega_3 & \omega_2 \\ \omega_3 & 0 & -\omega_1 \\ -\omega_2 & \omega_1 & 0 \end{bmatrix} \quad (3)$$

$\mathbf{I}_b$ is the moment of inertia, which is usually regarded as a diagonal matrix:

$$\mathbf{I}_b = \mathrm{diag}(\mathrm{I}_b^1 \quad \mathrm{I}_b^2 \quad \mathrm{I}_b^3) \quad (4)$$

Every rotor can generate aerodynamic thrust $F = \mathrm{c}_t u_j^2$ and aerodynamic torque $k_m u_j^2$. The models of resultant thrust $\mathbf{T}_b$ and torque $\mathbf{M}_b$ are as follows:

$$\begin{aligned}\mathbf{T}_b &= \sum_{j=1}^{\mathrm{K}} \mathbf{e}_3 F_j^b = \sum_{j=1}^{\mathrm{K}} \mathrm{c}_t u_j^2 \mathbf{e}_3 \\ \mathbf{M}_b &= \sum_{j=1}^{\mathrm{K}} \left( \mathbf{p}_{r_j}^b \times \mathbf{e}_3 F_j^b + (-1)^{j-1} \mathrm{k}_m \mathbf{e}_3 u_j^2 \right)\end{aligned} \quad (5)$$

where $u_j$ is the $j$-th ( $1 \le j \le \mathrm{K}$) rotor's angular speed, $\mathrm{c}_t$ is the rotor aerodynamic thrust coefficient, and $\mathrm{k}_m$ is the rotor aerodynamic torque coefficient. $\mathbf{p}_{r_j}^b$ is the $j$-th rotor's position in the body frame.

Using quadcopter as an example, as shown in Fig 3, the typical quadrotor's thrust and torque model based on (5) is as follows:

$$\mathrm{g}(\mathbf{u}) = \begin{bmatrix} 0 & 0 & 0 & 0 \\ 0 & 0 & 0 & 0 \\ \mathrm{c}_t & \mathrm{c}_t & \mathrm{c}_t & \mathrm{c}_t \\ -\mathrm{c}_t \mathrm{l}_y & \mathrm{c}_t \mathrm{l}_y & \mathrm{c}_t \mathrm{l}_y & -\mathrm{c}_t \mathrm{l}_y \\ -\mathrm{c}_t \mathrm{l}_x & -\mathrm{c}_t \mathrm{l}_x & \mathrm{c}_t \mathrm{l}_x & \mathrm{c}_t \mathrm{l}_x \\ \mathrm{k}_m & -\mathrm{k}_m & \mathrm{k}_m & -\mathrm{k}_m \end{bmatrix} \begin{bmatrix} u_1^2 \\ u_2^2 \\ u_3^2 \\ u_4^2 \end{bmatrix} \quad (6)$$

where $\mathrm{l}_x$ and $\mathrm{l}_y$ denote the projected lengths of the arm to the body center along the x-axis and y-axis, respectively, in the body coordinate system.

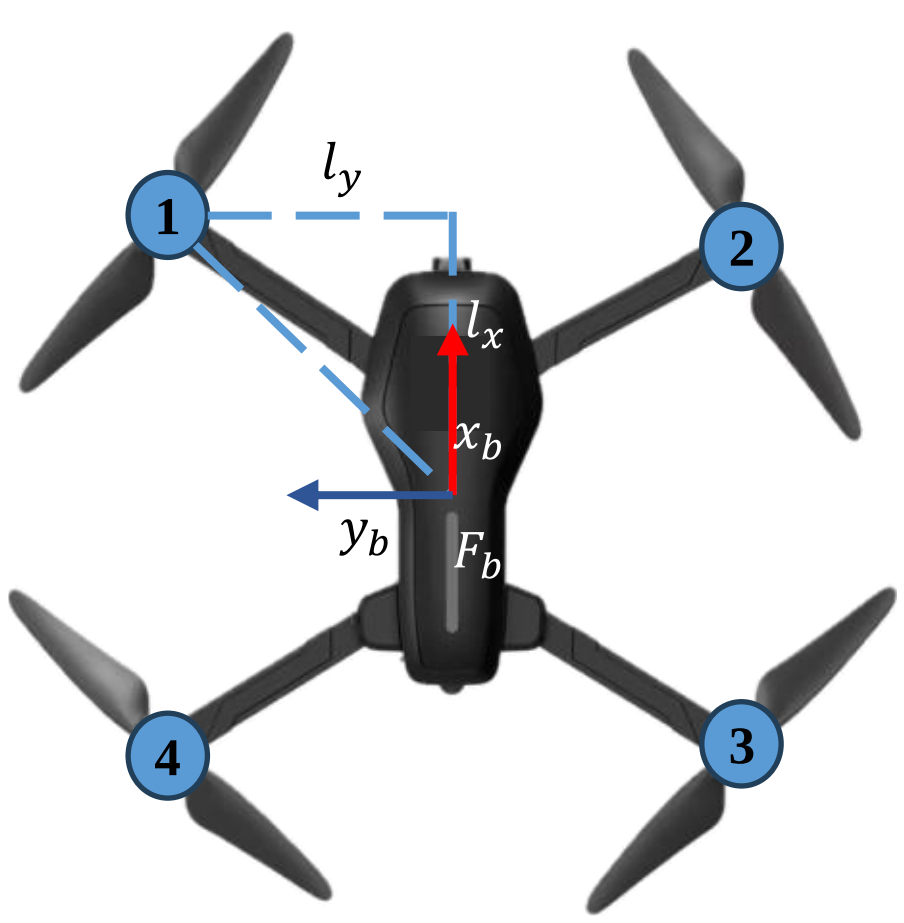

Fig. 3. The structure of the quadrotor.

To determine the covariance matrix of dynamic factors, an explicit noise model with additive Gaussian white noise is assumed as follows:

$$F_j = \mathrm{c}_t u_j^2 + n_A, n_A \sim \mathcal{N}(0, s) \quad (7)$$

where $s$ is the variance. Based on the noise model assumption, the thrust covariance matrix $\mathbf{P}_{\mathrm{T}}^i$ and torque covariance matrix $\mathbf{P}_{\mathrm{M}}^i$ of the quadrotor can be derived as follows:

$$\mathbf{P}_{\mathrm{T}}^i = 4 \begin{bmatrix} 0 & 0 & 0 \\ 0 & 0 & 0 \\ 0 & 0 & s \end{bmatrix} \quad (8a)$$

$$\mathbf{P}_{\mathrm{M}}^i = 4 \begin{bmatrix} \mathrm{l}_y^2 s & 0 & 0 \\ 0 & \mathrm{l}_x^2 s & 0 \\ 0 & 0 & \mathrm{k}_m^2 s \end{bmatrix} \quad (8b)$$

*B. Positioning based on FGO*

For simplicity, we assume that the observations of the absolute factor include position, attitude, velocity, and angular velocity in the simulation. We use multivariate Gaussian measurement assumption for state observation [31]:

$$\Phi_{\mathbf{x}} \propto \exp\left\{ \frac{1}{2} \| \mathrm{z}(\mathbf{x}_c) - \mathbf{z}_c \|_{\mathbf{P}_c}^2 \right\} \quad (9)$$

where $\mathbf{z}_c$ is the actual observation, and $\mathrm{z}(\mathbf{x}_c)$ is the predicted observation model.

In addition, the relative pose transformation $\mathbf{T}_{l+1}^l$ from the timestamp $l$ to the timestamp $l+1$ is denoted as follows:

$$\mathbf{T}_{l+1}^l = \left( \mathbf{T}_{b_l}^w \right)^{-1} \mathbf{T}_{b_{l+1}}^w \quad (10)$$

where $\mathbf{T}_{b_l}^w$ represents the UAV's pose at the timestamp $l$, $\mathbf{T}_{b_{l+1}}^w$ represent the UAV's pose at the timestamp $l+1$.

Hence, the relative factor is simplified as:

$$\Phi_{\mathbf{L}} \propto \exp\left\{ \frac{1}{2} \left\| \mathrm{Logmap}\left\{ \mathbf{T}_{l+1}^l \left( \mathbf{T}_{b_{l+1}}^w \right)^{\mathrm{T}} \mathbf{T}_{b_l}^w \right\} \right\|_{\mathbf{P}_l^L}^2 \right\} \quad (11)$$

where $\mathbf{P}_l^L$ is the covariance matrix of a relative pose from timestamp $l$ to timestamp $l+1$.

Based on (9) and (11), the residuals of absolute factor $\mathbf{r}_c^A$ and relative factor $\mathbf{r}_l^R$ are denoted as follows, respectively:

$$\begin{aligned}\mathbf{r}_c^A &= z(\mathbf{x}_c) - \mathbf{z}_c \\ \mathbf{r}_l^R &= \mathrm{Logmap}\left\{ \mathbf{T}_{l+1}^l \left( \mathbf{T}_{b_{l+1}}^w \right)^{\mathrm{T}} \mathbf{T}_{b_l}^w \right\}\end{aligned} \quad (12)$$

*C. MPC based on FGO*

The general MPC problem [4-6] is minimize a quadratic penalty on the error between the predicted states and inputs and a given dynamically feasible reference $\mathbf{z}_k^r$ and $\mathbf{u}_k^r$ for the future time horizon $N$. Consequently, the cost function is as follows:

$$\begin{aligned}&\min_{\mathbf{u}_{0:N-1}} \sum_{k=0}^{N-1} \left( \mathbf{x}_k^{\mathrm{T}} \mathbf{W}_k \mathbf{z}_k^r + \mathbf{u}_k^{\mathrm{T}} \mathbf{Y}_k \mathbf{u}_k^r \right) + \mathbf{x}_N^{\mathrm{T}} \mathbf{W}_N \mathbf{z}_N^r, \\ &s.t. \mathbf{x}_{k+1} = \mathrm{F}(\mathbf{x}_k, \mathbf{u}_k), 0 \le k \le N-1, \\ &\mathbf{x}_0 = \mathbf{x}_{init}, \\ &\mathbf{u}_{min} \le \mathbf{u}_k \le \mathbf{u}_{max},\end{aligned} \quad (13)$$

where $\mathbf{W}_k \ge 0$, $\mathbf{R} > 0$ and $\mathbf{W}_N \ge 0$ are the state, input, and final state weighting matrices, respectively. N is the $\mathbf{u}_k$ and $\mathbf{u}_k^r$ $(0 \le k \le N-1)$ are the input vector and the input set-point vector, respectively. $\mathbf{z}_k^r$ $(1 \le k \le N-1)$ and $\mathbf{z}_N^r$ are reference states.

The control input's change rate $\mathbf{u}_t - \mathbf{u}_{t+1}$ is restricted due to its driving capability. Furthermore, the input adheres to the bound limit function $\mathrm{h}(\mathbf{u})$. Further details will be discussed in Section III. Consequently, (13) can be reformulated for representation in a factor graph as follows:

$$\min_{u_{0:N-1}} \left\{ \begin{array}{c} \sum_{k=1}^{N-1} \|\mathbf{x}_k - \mathbf{z}_k^r\|_{\mathbf{Q}_k}^2 + \|\mathbf{x}_N - \mathbf{x}_N^r\|_{\mathbf{Q}_N}^2 \\ + \sum_{t=0}^{N-2} \|\mathbf{u}_t - \mathbf{u}_{t+1}\|_{\mathbf{R}_t}^2 + \sum_{j=0}^{N-1} \left\|\mathrm{h}(\mathbf{u}_j)\right\|_{\mathbf{Q}_{\mathrm{lim}}}^2 \end{array} \right\}, \tag{14}$$

$$s.t.\ \mathbf{x}_{k+1} = \mathrm{F}(\mathbf{x}_k, \mathbf{u}_k), 0 \le k \le N-1, \mathbf{x}_0 = \mathbf{x}_{init}$$

where $\mathbf{Q}_k$, $\mathbf{Q}_N$, $\mathbf{R}_t$, and $\mathbf{Q}_{\mathrm{lim}}$ represent different weighting matrices.

## III. Methodology

We combine positioning and control into a quadratic problem. Assuming $M$ is the number of states to be estimated and $N$ is the length of the predicted states. The unknown variables are $\boldsymbol{\chi} = \{\mathbf{x}_{-M+1:N}^{\mathrm{T}}, \mathbf{u}_{0:N-1}^{\mathrm{T}}\}^{\mathrm{T}}$. The unified optimization problem combines absolute factors, relative factors, and control-related factors. Therefore, JPCM's optimization problem is as follows:

$$\min_{\boldsymbol{\chi}} \left\{ \begin{array}{c} \sum_{c=-M+1}^{0} \|\mathbf{r}_c^A\|_{\mathbf{P}_c^A}^2 + \sum_{l=-M+1}^{-1} \|\mathbf{r}_l^R\|_{\mathbf{P}_l^L}^2 + \sum_{i=0}^{N-1} \|\mathbf{r}_i^D\|_{\mathbf{P}_i^D}^2 \\ + \sum_{k=1}^{N} \left\|\mathbf{r}_k^{ref}\right\|_{\mathbf{Q}_k}^2 + \sum_{t=0}^{N-2} \|\mathbf{r}_t^U\|_{\mathbf{R}_t}^2 + \sum_{j=0}^{N-1} \left\|\mathrm{h}(\mathbf{u}_j)\right\|_{\mathbf{Q}_{\mathrm{lim}}}^2 \end{array} \right\} \tag{15}$$

Additionally, the control part involves the reference trajectory factor residuals $\mathbf{r}^{ref} = \mathbf{x} - \mathbf{x}^{ref}$, dynamic factor residuals $\mathbf{r}_i^D = \mathbf{x}_{i+1} - \mathrm{F}(\mathbf{x}_i, \mathbf{u}_i)$, control bound factor residuals $\mathbf{h}(\mathbf{u}_j)$, and control rate factor residuals $\mathbf{r}_t^U = \mathbf{u}_t - \mathbf{u}_{t+1}$. These factors are further explained in the following sections.

To simplify the cost function of (15), we obtain:

$$\min_{\chi}\{\mathrm{f}_{\mathrm{sp}} + \|\mathrm{H}_0\mathrm{x}_0 - \mathrm{z}_0\|_{\mathrm{P}_0}^2 + \mathrm{f}_{\mathrm{dyn}} + \mathrm{f}_{\mathrm{MPC}}\} \tag{16}$$

where $\mathrm{f}_{dyn} = \sum_{i=0}^{N-1} \|\mathbf{r}_i^D\|_{\mathbf{P}_i^D}^2$ is the dynamic-related cost term, $\mathbf{H}_0$ is the Jacobian matrix at $\hat{\mathbf{x}}_0$, and $\mathbf{z}_0$ is the observation of the initial state, and $\mathrm{f}_{MPC}$ is the cost function of MPC. Absolute factors and relative factors are reformulated to express the residuals $\mathrm{H}(\mathbf{x}_0) - \mathbf{z}_0$ for the initial state $\mathbf{x}_0$ and the cost function term $\mathrm{f}_{sp}$ for the subsequent states. If only the current state $\mathbf{x}_0$ is considered, $\mathrm{f}_{sp}$ equals zero.

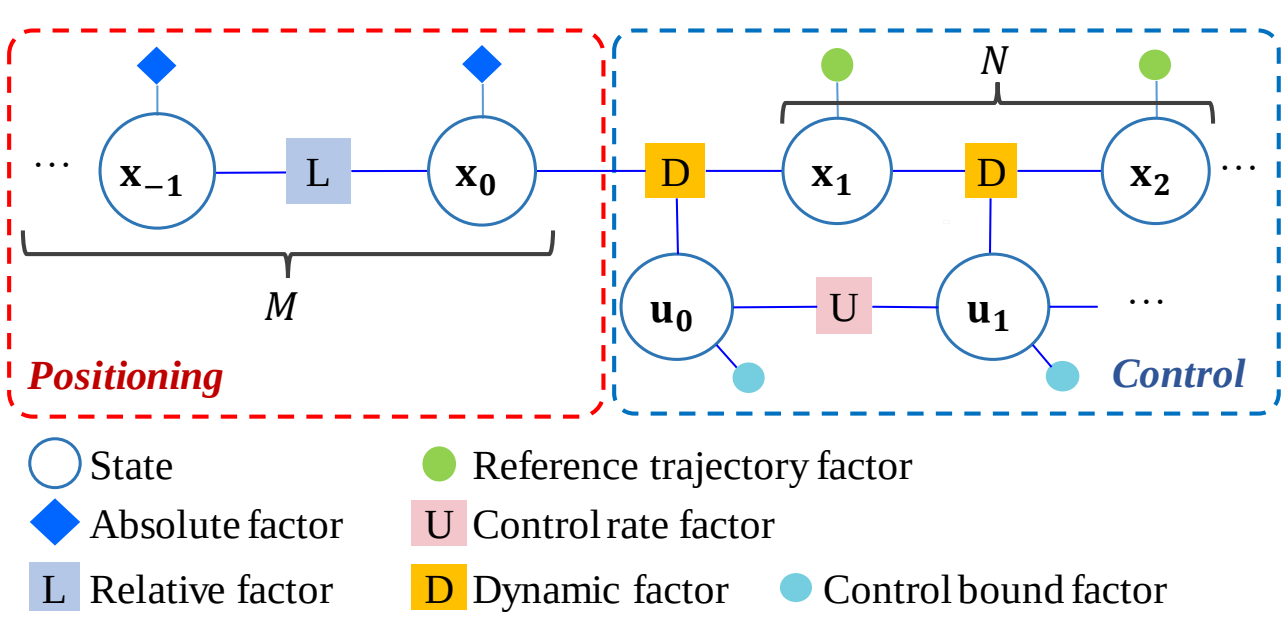

Fig. 4. The factor graph of the proposed JPCM. The absolute factor and relative factor are two types of factors generated from common sensors, respectively. The dynamic factor is derived from the UAV's dynamic model, which bridges the gap between positioning and the control module. The reference trajectory factor makes predicted states close to the pre-planned trajectory.

Therefore, we provide a unified factor graph framework to implement the proposed JPCM. As illustrated in Fig 4, this unified factor graph slides over a time window, which contains $M$ states and $N$ predicted states. The positioning component, highlighted in red, incorporates both absolute and relative factors. In our simulation, $N = 20$, and $M$ depends on the size of the sliding window. Additionally, the control component comprises the dynamic factor, reference trajectory factor, control rate factor, and control bound factor.

### A. Dynamic factor

The dynamic factor serves as a crucial link, bridging the gap between positioning and control. The discrete form of state propagation is $\mathbf{x}_{i+1} = \mathbf{x}_i + \dot{\mathbf{x}}_i\Delta\mathrm{t}$. Thus, ignoring aerodynamic drag, the dynamic error can be represented through $\mathbf{r}_i^D = [\mathbf{e}_p^{\mathrm{T}}, \mathbf{e}_\theta^{\mathrm{T}}, \mathbf{e}_v^{\mathrm{T}}, \mathbf{e}_\omega^{\mathrm{T}}]^{\mathrm{T}} = \mathbf{x}_{i+1} - \mathbf{x}_i - \dot{\mathbf{x}}_i\Delta\mathrm{t}$ as follows:

$$\begin{aligned} \mathbf{e}_{\mathrm{p}} &= \mathbf{p}_{b_{i+1}}^w - \mathbf{v}_{b_i}^w * \Delta\mathrm{t} - \mathbf{p}_{b_i}^w \\ \mathbf{e}_\theta &= \mathrm{Log}(\mathbf{R}_w^{b_{i+1}}\mathbf{R}_{b_i}^w(\mathbf{I}_{3\times3} + \boldsymbol{\omega}_b^\times\Delta\mathrm{t}) \\ \boldsymbol{e}_{\mathrm{v}} &= \mathbf{v}_{b_{i+1}}^w - \mathbf{v}_{b_i}^w - \left(-\mathbf{R}_G^w\mathbf{e}_3\mathrm{g} + \mathbf{R}_{b_i}^w\mathbf{T}_{b_i}/\mathrm{m}_\mathrm{b}\right)\cdot\Delta\mathrm{t} \\ \mathbf{e}_\omega &= \boldsymbol{\omega}_{b_{i+1}} - \boldsymbol{\omega}_{b_i} - \mathbf{I}_\mathrm{b}^{-1}(\mathbf{M}_{b_i} - \boldsymbol{\omega}_{b_i} \times \mathbf{I}_\mathrm{b}\boldsymbol{\omega}_{b_i})\cdot\Delta\mathrm{t} \end{aligned} \tag{17}$$

where $\Delta\mathrm{t}$ is the period between two contiguous states. $\mathrm{Log}(*)$ and $\mathrm{Exp}(*)$ represent and Logarithmic map and Exponential map of Lie theory, respectively.

To propagate covariance, the dynamic model factor's error function $\mathbf{r}_i^D = [\bar{\mathbf{e}}_p^{\mathrm{T}}, \bar{\mathbf{e}}_\theta^{\mathrm{T}}, \bar{\mathbf{e}}_v^{\mathrm{T}}, \bar{\mathbf{e}}_\omega^{\mathrm{T}}]^{\mathrm{T}}$ can be derived from the error model (18):

$$\begin{aligned} \bar{\mathbf{e}}_p &= \mathbf{R}_w^{b_i}\mathrm{m}_b\big(\mathbf{p}_{b_{i+1}}^w - \mathbf{v}_{b_i}^w\Delta\mathrm{t} + 0.5\mathbf{R}_G^w\mathbf{e}_3\mathrm{g}\Delta\mathrm{t}^2 - \mathbf{p}_{b_i}^w\big) \\ &\quad -0.5\mathbf{T}_{b_i}\Delta\mathrm{t}^2 \\ \bar{\mathbf{e}}_\theta &= \mathrm{Log}\big(\mathbf{R}_w^{b_i}\mathbf{R}_{b_{i+1}}^w\big) - \boldsymbol{\omega}_{b_i}\Delta\mathrm{t} \\ \bar{\boldsymbol{e}}_v &= \mathbf{R}_w^{b_i}\mathrm{m}_b\big(\mathbf{v}_{b_{i+1}}^w - \mathbf{v}_{b_i}^w + \mathbf{R}_G^w\mathbf{e}_3\mathrm{g}\Delta\mathrm{t}\big) - \mathbf{T}_{b_i}\Delta\mathrm{t} \\ \bar{\mathbf{e}}_\omega &= \mathbf{I}_b\big(\boldsymbol{\omega}_{b_{i+1}} - \boldsymbol{\omega}_{b_i}\big) - (\mathbf{M}_{b_i} - \boldsymbol{\omega}_{b_i} \times \mathbf{I}_\mathrm{b}\boldsymbol{\omega}_{b_i})\cdot\Delta\mathrm{t} \end{aligned} \tag{18a}$$

If aerodynamic drag is further considered, according to (2), the velocity component $\bar{\mathbf{e}}_v$ of the dynamic factor's residual is as follows:

$$\begin{aligned} \bar{\mathbf{e}}_v = \mathbf{R}_w^{b_i}\mathrm{m}_b\big(\mathbf{v}_{b_{i+1}}^w - \mathbf{v}_{b_i}^w + \mathbf{R}_G^w\mathbf{e}_3\mathrm{g}\Delta\mathrm{t}\big) - \mathbf{T}_{b_i}\Delta\mathrm{t} \\ - \mathbf{D}\mathbf{R}_w^{b_i}\mathbf{v}_{b_i}^w\Delta\mathrm{t} \end{aligned} \tag{18b}$$

According to (8) and (18a), the covariance matrix of $\bar{\mathbf{e}}_v$ and $\bar{\mathbf{e}}_\omega$ are $\mathbf{P}_\mathrm{T}^i\Delta\mathrm{t}^2$ and $\mathbf{P}_\mathrm{M}^i\Delta\mathrm{t}^2$, respectively. Moreover, the covariance matrix of $\bar{\mathbf{e}}_p$ is set by $0.25\mathbf{P}_T^i\Delta\mathrm{t}^4$. The Jacobian matrix is provided in the Appendix.

### B. Control bound factor and control rate factor

The control bound factor is designed to ensure that the control input meets the actuator's limits. Defining the hinge loss cost function $\mathbf{E} = \mathrm{h}(\mathbf{u}_j)$ for the control input inequality, which is as follows:

$$\mathbf{E}_i = \begin{cases} u_{min}^j + u_{thr}^j - \mathbf{u}_j(i) & \text{if } \mathbf{u}_j(i) < u_{min}^j + u_{thr}^j \\ \mathbf{u}_j(i) - u_{max}^j + u_{thr}^j & \text{if } u_{max}^j - u_{thr}^j \le \mathbf{u}_j(i) \\ 0 & \text{otherwise} \end{cases} \quad (1 \le i \le K) \tag{19}$$

where $u_{min}^j$ is the input lower bound, and $u_{max}^j$ is the input upper bound, and $u_{thr}^j$ is the threshold value. In addition, $\mathbf{Q}_{lim}$ is a control bound factor's weighting matrix which determines how fast the error grows as the value approaches the limit.

To smooth the trajectory, the control rate factor is used to limit the rate of control inputs, which is as follows:

$$\mathbf{r}_t^U = \mathbf{u}_t - \mathbf{u}_{t+1} \tag{20}$$

*C. Reference trajectory factor*

The reference trajectory factor's residual is as follows:

$$\mathbf{r}^{ref} = (\mathbf{p}, \mathbf{R}, \mathbf{v}) \ominus (\mathbf{p}^r, \mathbf{R}^r, \mathbf{v}^r) \tag{21}$$

where $\ominus$ is a subtraction operator, which calculates the residuals of position, velocity, and rotation. The position residual is $\mathbf{e}_p^{ref} = \mathbf{p} - \mathbf{p}^r$, the velocity residual is $\mathbf{e}_v^{ref} = \mathbf{v} - \mathbf{v}^r$, and the rotation residual is $\mathbf{e}_\theta^{ref} = \mathrm{Log}(\mathbf{R}^{\mathrm{T}}\mathbf{R}^r)$. Then, $\mathbf{r}^{ref} = \left[\left(\mathbf{e}_p^{ref}\right)^{\mathrm{T}} \quad \left(\mathbf{e}_v^{ref}\right)^{\mathrm{T}} \quad \left(\mathbf{e}_\theta^{ref}\right)^{\mathrm{T}}\right]^{\mathrm{T}}$. In addition, the reference trajectory factor's covariance $\mathbf{Q}_k (1 \le k \le N)$ can be configured by users.

## IV. Numerical Simulation and Experiments

*A. Simulation scheme*

The quadrotor's trajectory control simulation is conducted to evaluate the proposed JPCM's performance. The reference path is assumed to be a circle. Then, the trajectory is generated by the minimum snap trajectory generation method [43]. Moreover, the Gaussian white noise is added to the thrust and torque of the quadrotor simulator. The full implementation of the model in this paper is open-source to benefit the research community.

The gravity of the quadrotor is 10 N (Newton). The thrust noise's sigma is 0.1N, and the angular speed process noise's sigma is 0.02rad/s. The simulated circle trajectory's linear speed is 5m/s, and its radius is 1.5m. The control frequency is 100 Hz. We assume the absolute factor's measurements include position, velocity, angular speed, and attitude, all with simulated uncertainty. As shown in Fig 5, we simulated a quadrotor based on the quadratic thrust model (5). In our dynamic model, we incorporate various disturbances including the aerodynamic drag force, Gaussian noise, and abrupt motion. This enables us to validate the trajectory tracking performance of our proposed algorithm under challenging conditions characterized by high uncertainty.

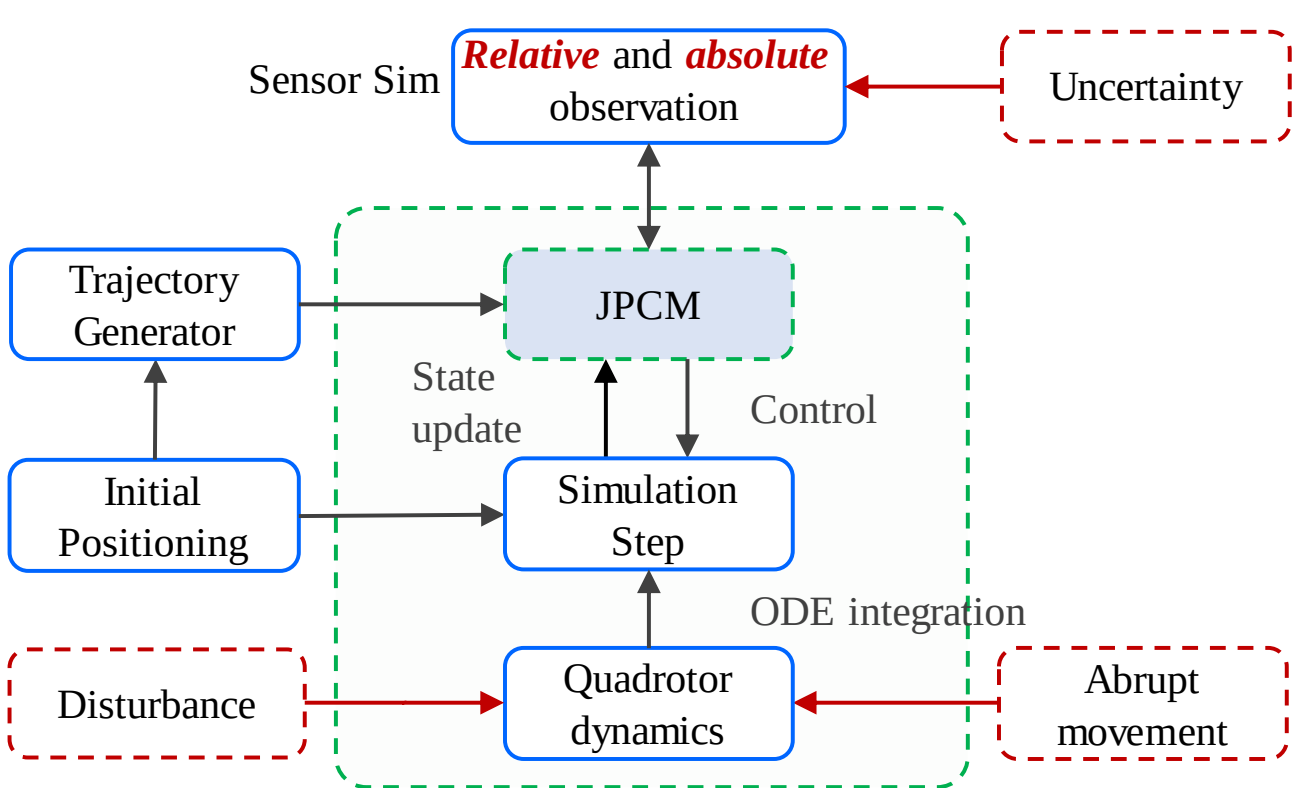

Fig. 5. Simulation framework.

*B. Experiment scheme*

Lemniscate trajectory tracking experiments are conducted to demonstrate the proposed JPCM method. The data from the motion capture system is used as the ground truth (position, rotation, velocity, and angular velocity) and after adding Gaussian noise, it is used to simulate GNSS position measurements. In addition, a lost-cost IMU is also used to collect acceleration and angular velocity. The quadrotor and its system framework are illustrated in Fig 6 (a) and Fig 6 (b), respectively. The performance of the popular motion control model (time division of positioning and control) is compared with the proposed JPCM. In the traditional model, the IMU and GNSS factors [12] are fused using FGO in the positioning. In addition, thrust $\mathbf{T}_{des}$ and angular velocity $\boldsymbol{\omega}_{des}$ are selected as the control inputs of the MPC and proposed JPCM. Then, a low-level controller embedded in Pixhawk is adopted to track the desired thrust and angular velocity [5]. In addition, the voltage and current can be recorded to compute the power.

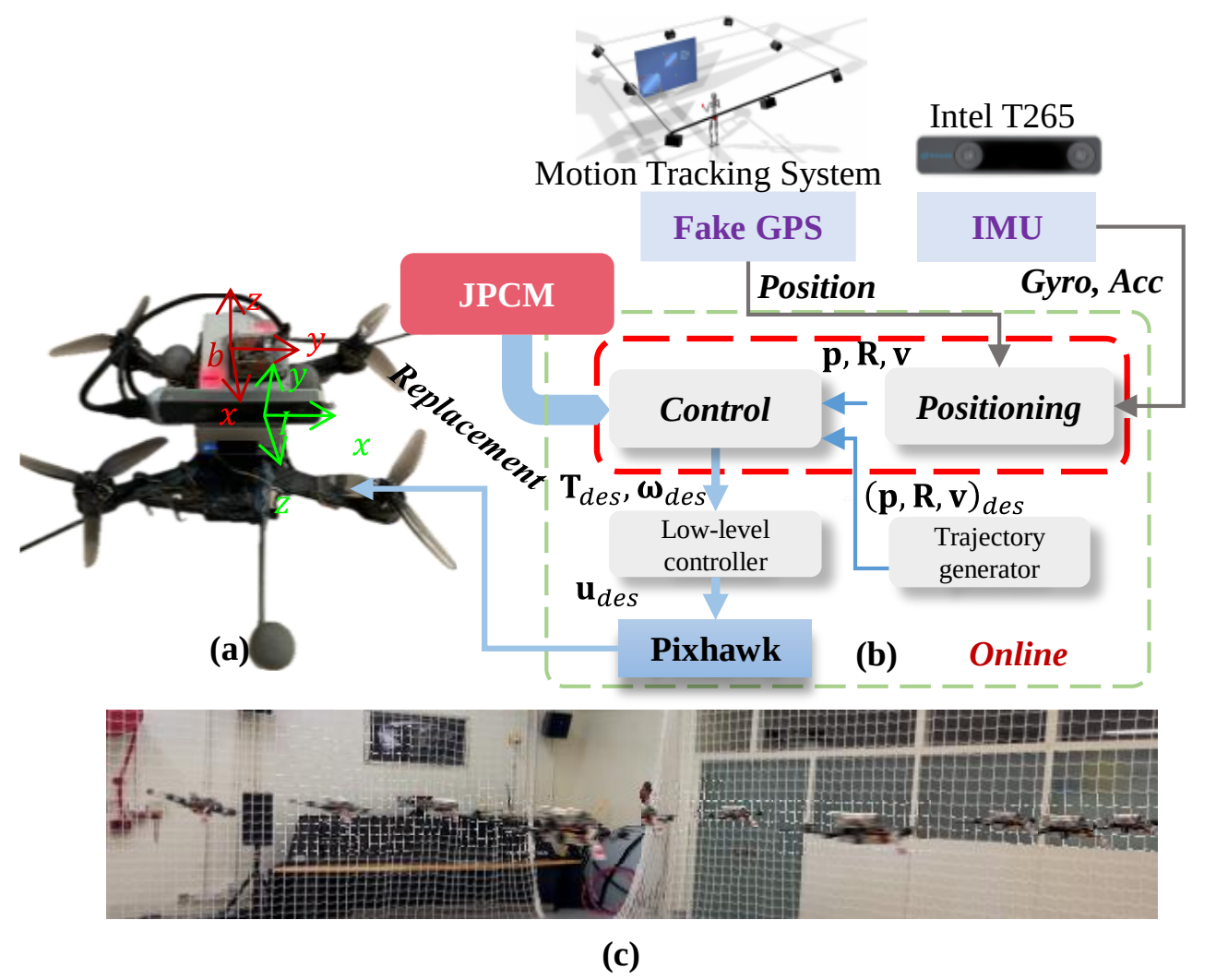

Fig. 6. Experimental platform and system framework. (a) The quadrotor is used in experiments. (b) The framework of the quadrotor's automation system. (c) Flight experiment images.

The joined factor graph to demonstrate JPCM is shown in Fig 7 (a), which includes GNSS factors [12], IMU factors [3], and MPC-derived factors in Section III. First, the proposed JPCM is independent of sensor type by incorporating appropriate measurement-based constraints, sensors such as LiDAR [44], camera, GNSS, and IMU can all be seamlessly integrated. Second, complementary sensors are essential in urban environments. For instance, LiDAR has high local positioning accuracy. However, due to the non-line-of-sight effect of the satellite signals, only GNSS is used and the uncertainty is high. Although the proposed method has a more robust adaptive ability to variable positioning uncertainty. To ensure better trajectory tracking accuracy, sensors that complement both local and global perception should be utilized in urban environments.

Furthermore, to limit the computational load, the oddest state is marginalized once it falls outside the time window [45]. The pipeline of the proposed JPCM is illustrated in Fig 7 (b), and we separately cache the positioning factor graph for performing marginalization to construct the joined factor graph. For details on the JPCM's implementation can be found in our open-source code.

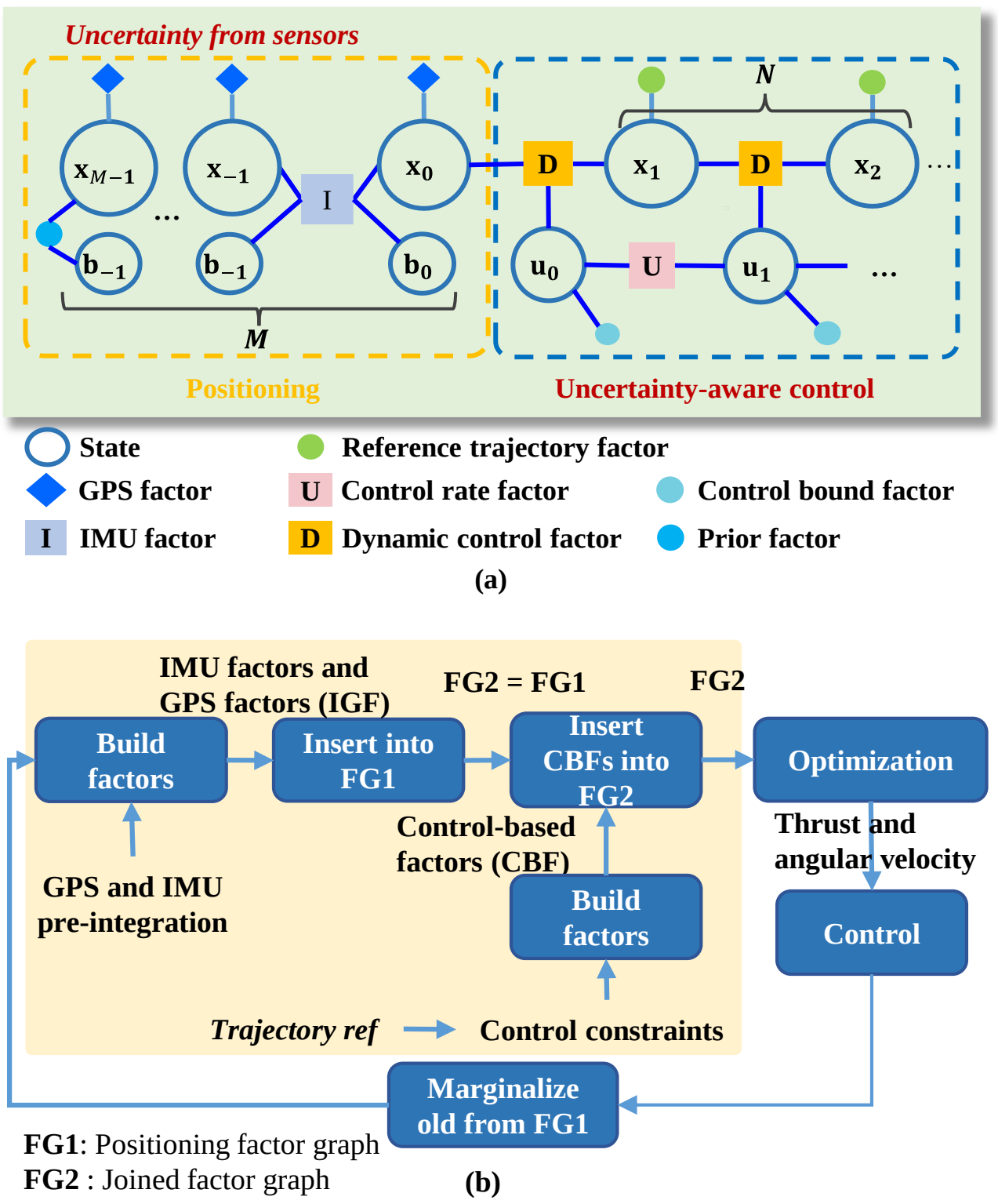

Fig. 7. Factor graph and its construction pipeline. (a) Factor graph of the proposed JPCM used in the experiments. (b) The pipeline of constructing the factor graph.

## C. Simulation results

*a) Performance comparison between MPC and JPCM with high positioning uncertainty*

*Illustration of the abbreviations of different methods to be evaluated in this paper:* (1) **MPC-pre** [8] represents the MPC method with highly precise positioning. (2) **MPC** [8] represents the MPC method with highly uncertain observation with parameters listed in Table **I.** (3) **JCPM** represents the proposed tightly joined positioning and control model with highly uncertain observation for Case 1. (4) **JCPM-SW** represents the proposed tightly joined positioning and control model based on a sliding window with highly uncertain observation for Case 2.

*Case 1:* First, it is assumed that only absolute measurements exist at the current moment. Similarly, a state with uncertainty output by the common positioning module is provided to the MPC and JPCM. Both MPC and JPCM are implemented based on FGO. MPC directly adopts the observation of the current state as the initial state.

*Case 2:* In addition to the conditions of case 1, it is assumed that the UAV is equipped with a LiDAR and can measure the pose transformation at adjacent timestamps. The length of the variables to be estimated is set to 10. The extrinsic parameter between the LiDAR frame and the body frame is an identity matrix. Verify the feasibility of the proposed method under these conditions. We simulate the relative pose transformation $\mathbf{T}_{l+1}^{l}$ with the Gaussian white noise $\mathbf{n}_l$ as follows:

$$\mathbf{T}_{l+1}^{l} = \widehat{\mathbf{T}}_{l+1}^{l}\mathrm{Expmap}(\mathbf{n}_l) \quad (22)$$

where $\widehat{\mathbf{T}}_{l+1}^{l}$ is the true value of relative pose.

All simulation parameters are listed in Table I.

TABLE I
Simulation parameters

| Parameters | Case 1 | Case 2 |
|---|---|---|
| $M$ (the length of historical states in positioning) | 1 | 10 |
| $N$ (the length of predicted states in control) | 20 | |
| Position sigma | 0.20 m | |
| Rotation sigma | 0.03 rad | |
| Velocity sigma | 0.05 m/s | |
| Angular velocity sigma | 0.001 rad/s | |
| Sigma of $\mathbf{T}_{l+1}^{l}$ | $\mathrm{diag}\{0.03^2\mathrm{I}_3^{\mathrm{T}}, 0.03^2\mathrm{I}_3^{\mathrm{T}}\}$ | |
| $\mathbf{P}_l^L$ | $\mathrm{diag}\{0.03^2\mathrm{I}_3^{\mathrm{T}}, 0.03^2\mathrm{I}_3^{\mathrm{T}}\}$ | |
| $\mathbf{Q}_k$ | $\mathrm{diag}\{0.03^2\mathrm{I}_3^{\mathrm{T}}, 0.3^2\mathrm{I}_3^{\mathrm{T}}, 3^2\mathrm{I}_3^{\mathrm{T}}\}$ | |
| $\mathbf{Q}_N$ | $\mathrm{diag}\{0.01^2\mathrm{I}_3^{\mathrm{T}}, 0.3^2\mathrm{I}_3^{\mathrm{T}}, 3^2\mathrm{I}_3^{\mathrm{T}}\}$ | |
| $\mathbf{R}_t$ | $\mathrm{diag}\{1000\mathrm{I}_4^{\mathrm{T}}\}$ | |
| $u_{min}$ | 12000 | |
| $u_{thr}$ | 100 | |
| $u_{max}$ | 18000 | |

*1) Simulation Results of Case 1*

For case 1, Fig 8 illustrates the path of MPC and JPCM, which shows that JPCM aligns more closely with the reference path than the MPC method. In contrast, the nominal MPC exhibits significant fluctuation.

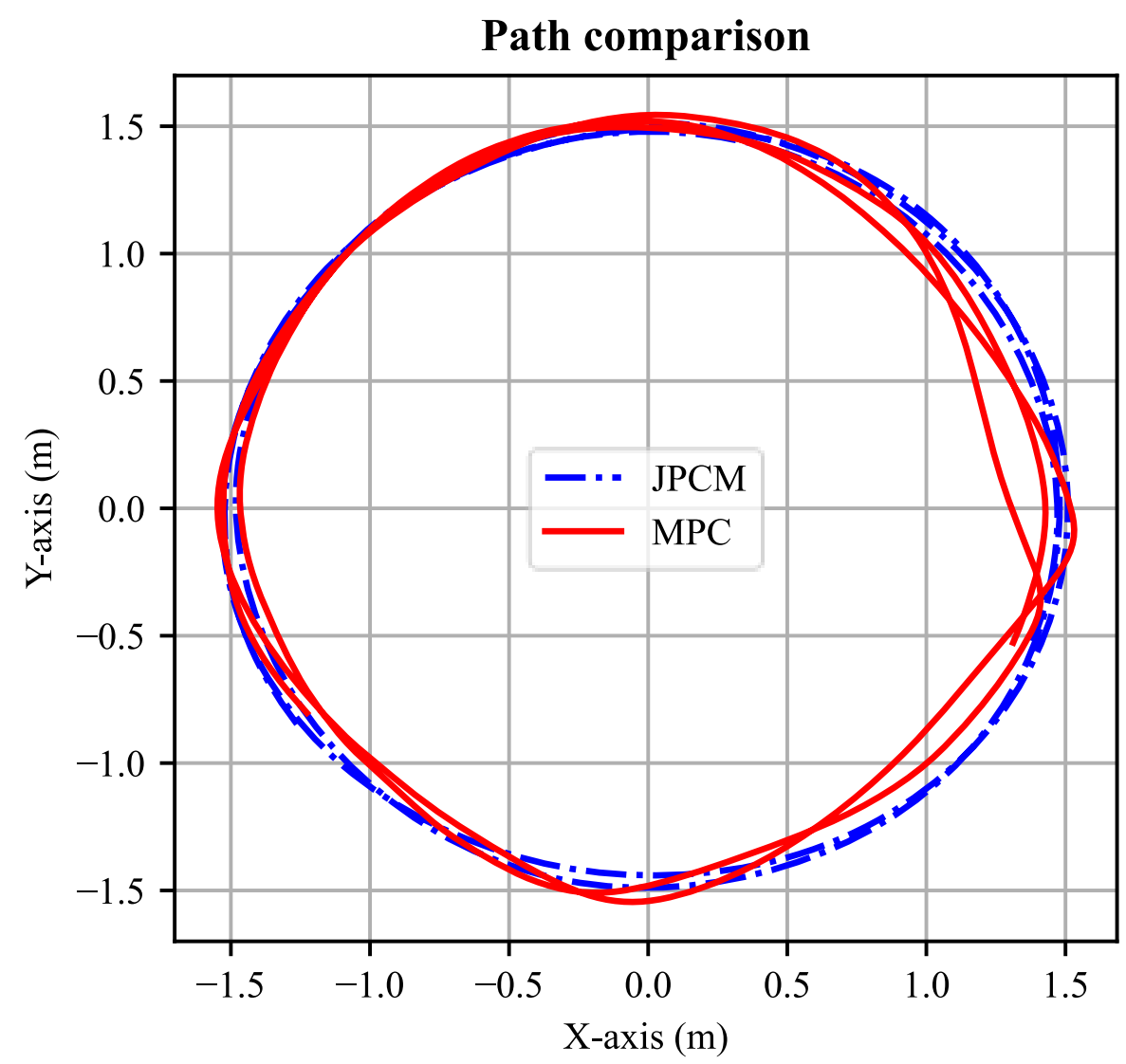

Fig. 8. Case 1: The paths of MPC and JPCM (linear speed = 5m/s, radius = 1.5m). The red solid line and blue solid line represent the tracking path based on MPC, and the tracking path based on JPCM, respectively.

The root mean square errors (RMSEs) are also summarized in Table II. MPC-pre represents the MPC method with high-precision positioning. When positioning measurements are precise, the MPC problem solved by FGO converges to the reference trajectory with minimal control errors. Table II reveals that both the position and rotation control errors of the proposed JPCM are reduced compared to the nominal MPC with large noise positioning. This indicates that noisy positioning adversely affects the control performance of UAVs.

TABLE II
THE TRAJECTORY FOLLOWING RMSE OF MPC-PRE (MPC WITH PRECISE POSITIONING), MPC, JPCM, AND JPCM-SW

| Method | Position RMSE (m) | | | Rotation RMSE (rad) | | |
|---|---|---|---|---|---|---|
| **MPC-pre** | 0.010 | 0.010 | 0.005 | 0.005 | 0.005 | 0.004 |
| **MPC** | 0.068 | 0.066 | 0.055 | 0.246 | 0.151 | 0.251 |
| **JPCM** | 0.031 | 0.019 | 0.015 | 0.031 | 0.018 | 0.158 |
| **JPCM-SW** | 0.017 | 0.017 | 0.019 | 0.010 | 0.011 | 0.205 |

As shown in Fig 9, under highly uncertain observations, the MPC's control inputs often reach the upper or lower bound. On the contrary, the control inputs of the JPCM are relatively concentrated. The MPC is probing at the edge of the control boundary, indicating that the solution of the control quantity is on the verge of reaching a critical constraint. Specifically, when the positioning deviates excessively from the target, the MPC problem may fail to find a feasible solution.

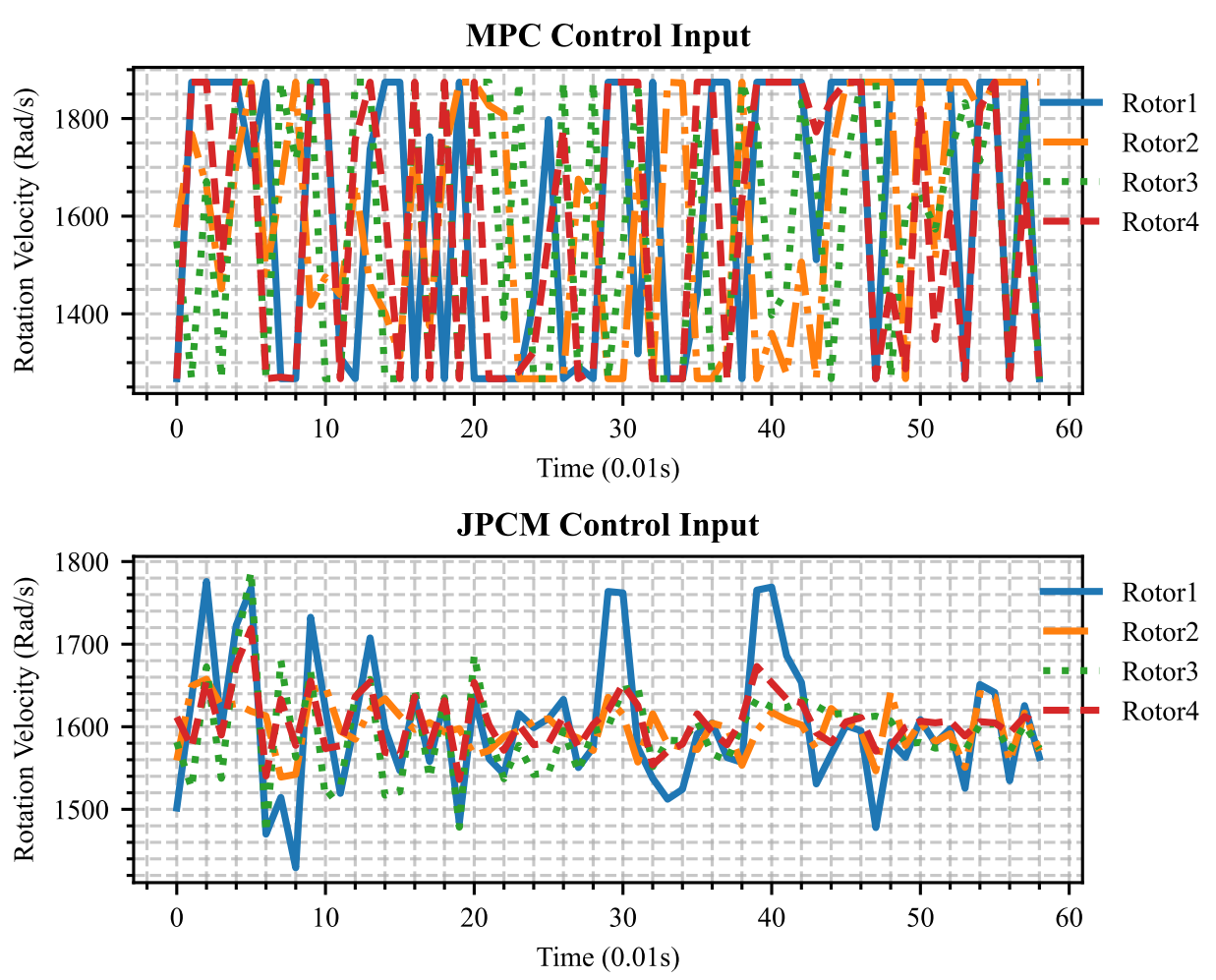

Fig. 9. Case 1: The comparison of control input between MPC and JPCM. Most control inputs are limited between upper and lower bounds.

2) *Simulation Results of Case 2*

As for case 2, the JPCM-SW is conducted with $M = 10$. As demonstrated in Table II, the JPCM-SW method exhibits a slight reduction in both the x-axis and y-axis position following residuals when compared to the standard JPCM method. The enhanced performance of JPCM-SW is achieved through a tightly integrated approach that effectively combines historical observations with control objectives. By utilizing more historical information, the JPCM-SW method achieves higher precise control than MPC.

MPC can handle most control tasks effectively when the positioning accuracy is high. However, in urban scenarios where severe noise is prevalent, the trajectory following the error of MPC may be significantly exacerbated. In contrast, the proposed JPCM substantially improves trajectory tracking performance in situations where positioning uncertainty is high. To conclude, the performance of JPCM based on FGO surpasses that of the standard MPC, particularly when precise positioning is not available.

*b) Performance comparison under disturbances*

1) *Abrupt movements*

Sudden winds or collision can cause UAVs to experience instantaneous movement. We refer to the recovery after a rapid and large movement as a "*recovery process*". We simulate a rapid and large movement at one moment to investigate the "*recovery process*".

As illustrated in Fig 10, an unplanned movement $\Delta\mathbf{p}_b^w = [0.00\text{m}, 0.30\text{m}, -0.40\text{m}]^\mathrm{T}$ is simulated at 0.5 seconds, and the UAV controlled by JPCM returns to the planned trajectory in about three seconds. In comparison to MPC, the recovery process of the proposed JPCM method takes a longer duration. However, JCPM renders a smoother position tracking error curve according to Fig 10. The disadvantage is that when JPCM experiences significant movement disturbances, the control delay will be greater. Conversely, in scenarios that pursue smoother control, this becomes an advantage. For example, in the case of positioning outliers or spoofing, smoother and slower JPCM is more robust than MPC. Besides, the control delay can be accepted in practical systems.

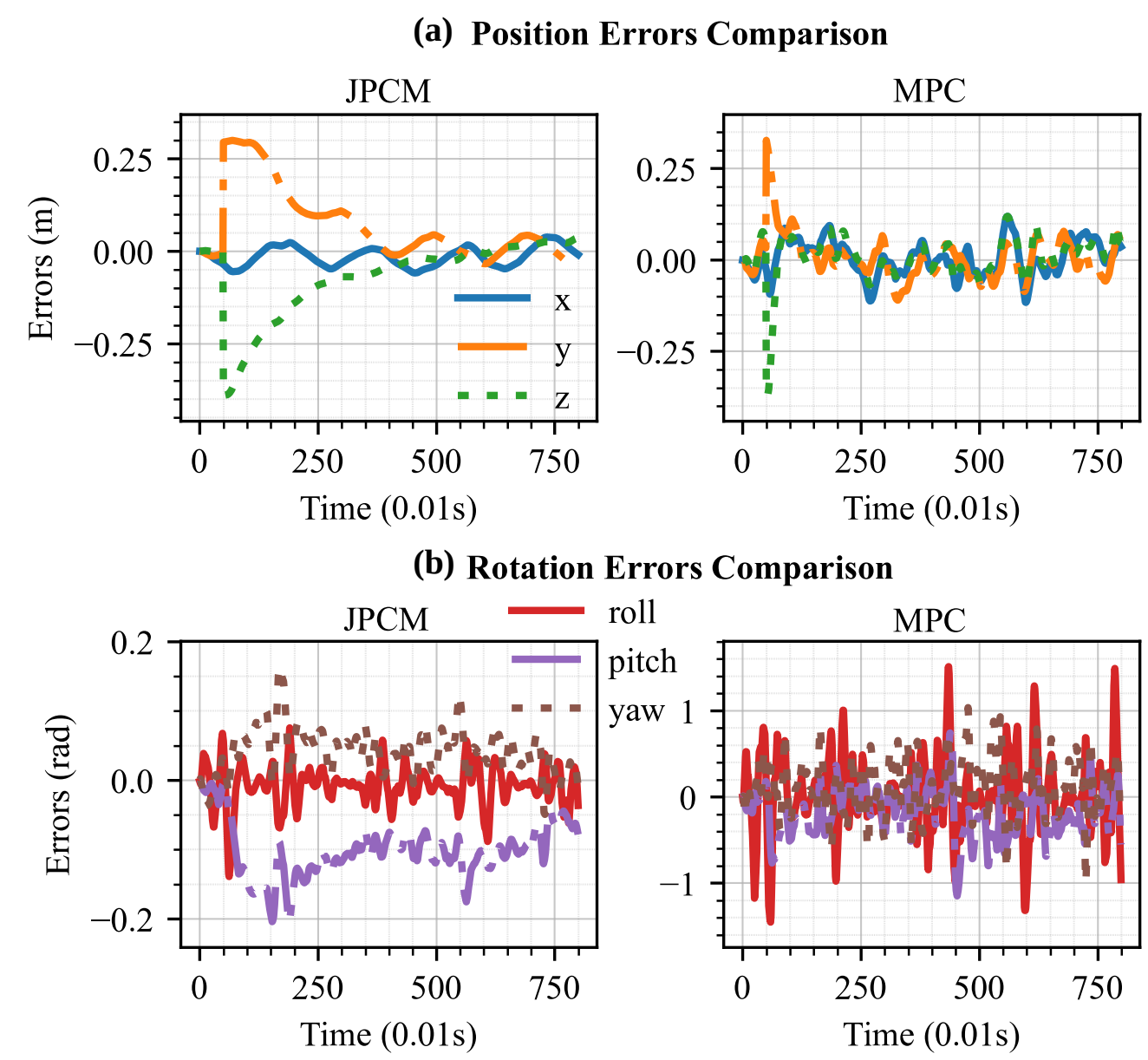

Fig. 10. Position tracking errors after an abrupt movement. (a) Position tracking errors comparison. (b) Rotation tracking errors comparison.

2) *Aerodynamic drag*

First, we investigate the performance of JPCM affected by the aerodynamic drag (not considering aerodynamic drag factors in the controller). Then, to mitigate the effects of aerodynamic drag force, we introduce a drag-aiding JPCM method, namely **JPCM-Drag**.

To evaluate the performance of the proposed JPCM under the simulated parameters in Table I Case 1, four prevalent aerodynamic drag coefficients are strategically chosen. An analysis of the data presented in Table III indicates a direct correlation between the trajectory position tracking error of JPCM and the magnitude of the aerodynamic drag coefficient.

TABLE III
THE TRAJECTORY FOLLOWING RMSE OF JPCM AND JPCM-DRAG WITH AERODYNAMIC DRAG FORCE

| Method | Drag parameter | Position RMSE (m) | | | Rotation RMSE (rad) | | |
|---|---|---|---|---|---|---|---|
| **JPCM** | **0** | 0.026 | 0.025 | 0.018 | 0.022 | 0.016 | 0.024 |
| **JPCM** | $\mathbf{0.1I}_{3\times3}$ | 0.040 | 0.031 | 0.024 | 0.023 | 0.013 | 0.045 |
| **JPCM** | $\mathbf{0.2I}_{3\times3}$ | 0.053 | 0.047 | 0.035 | 0.018 | 0.023 | 0.040 |
| **JPCM** | $\mathbf{0.3I}_{3\times3}$ | 0.067 | 0.061 | 0.039 | 0.032 | 0.036 | 0.062 |
| **JPCM-Drag** | $\mathbf{0.3I}_{3\times3}$ | 0.031 | 0.025 | 0.028 | 0.011 | 0.037 | 0.064 |

***Eliminating aerodynamic drag effects***: JPCM-Drag incorporates the aerodynamic drag into the error function (19b) and recalculates the Jacobian matrix of the modified dynamic model factor. This is achieved by integrating a non-zero coefficient matrix, denoted as $\mathbf{D}$, into the formulas provided in the Appendix. A specific instance of this method is demonstrated where $\mathbf{D}$ is set to $0.3\mathbf{I}_{3\times3}$ that is consistent with the actual system level. As shown in Fig 11, the JPCM method reveals a significant deviation in the circular trajectory towards the circle center. The shift is attributed to the influence of aerodynamic drag. The JPCM-Drag method, which accounts for this drag effect, tends to converge towards the pre-set trajectory. Furthermore, when the JPCM-Drag method is implemented, the RMSE of the position following closely approximates the RMSE in scenarios devoid of aerodynamic drag. This observation underscores the efficacy of the JPCM-Drag method in mitigating the impact of aerodynamic drag.

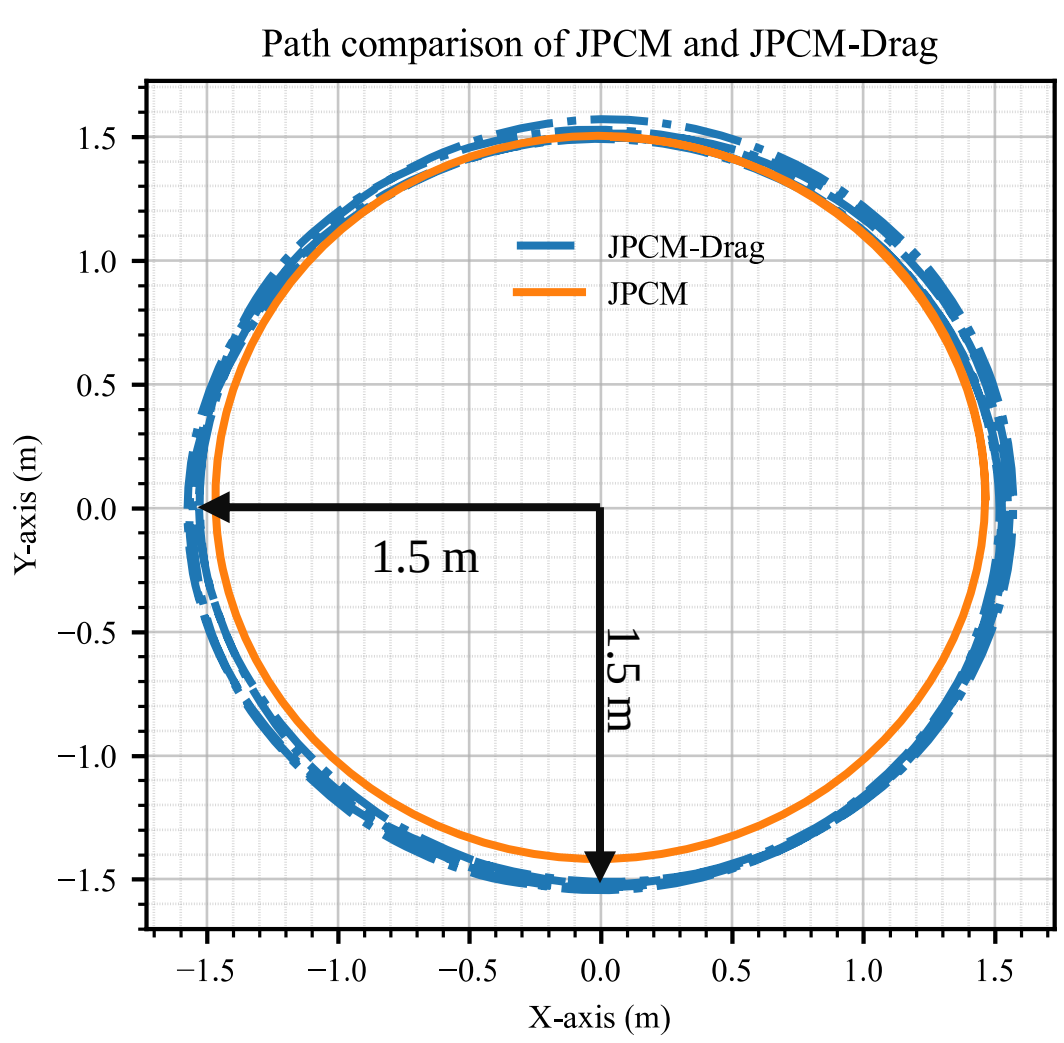

Fig. 11. Path comparison of JPCM and JPCM-Drag method with aerodynamic drag ($\mathbf{D} = 0.3\mathbf{I}_{3\times3}$).

*D. Experimental results and analysis*

*a) Tracking performance comparison*

This section shows trajectory tracking results of the proposed JPCM under precise positioning and large uncertainty scenarios. Besides, power is also discussed.

1) *MPC based on FGO (MPC-FGO) vs DFBC with precise positioning*

As shown in Fig 12, the MPC-FGO achieves RMSE values of 0.032 m, 0.043 m, and 0.0099 m on the x-axis, y-axis, and z-axis, respectively and shows smaller tracking error compared with the differential-flatness-based controller (DFBC). These results demonstrate path-tracking accuracy comparable to the state-of-the-art MPC method [5], given precise positioning.

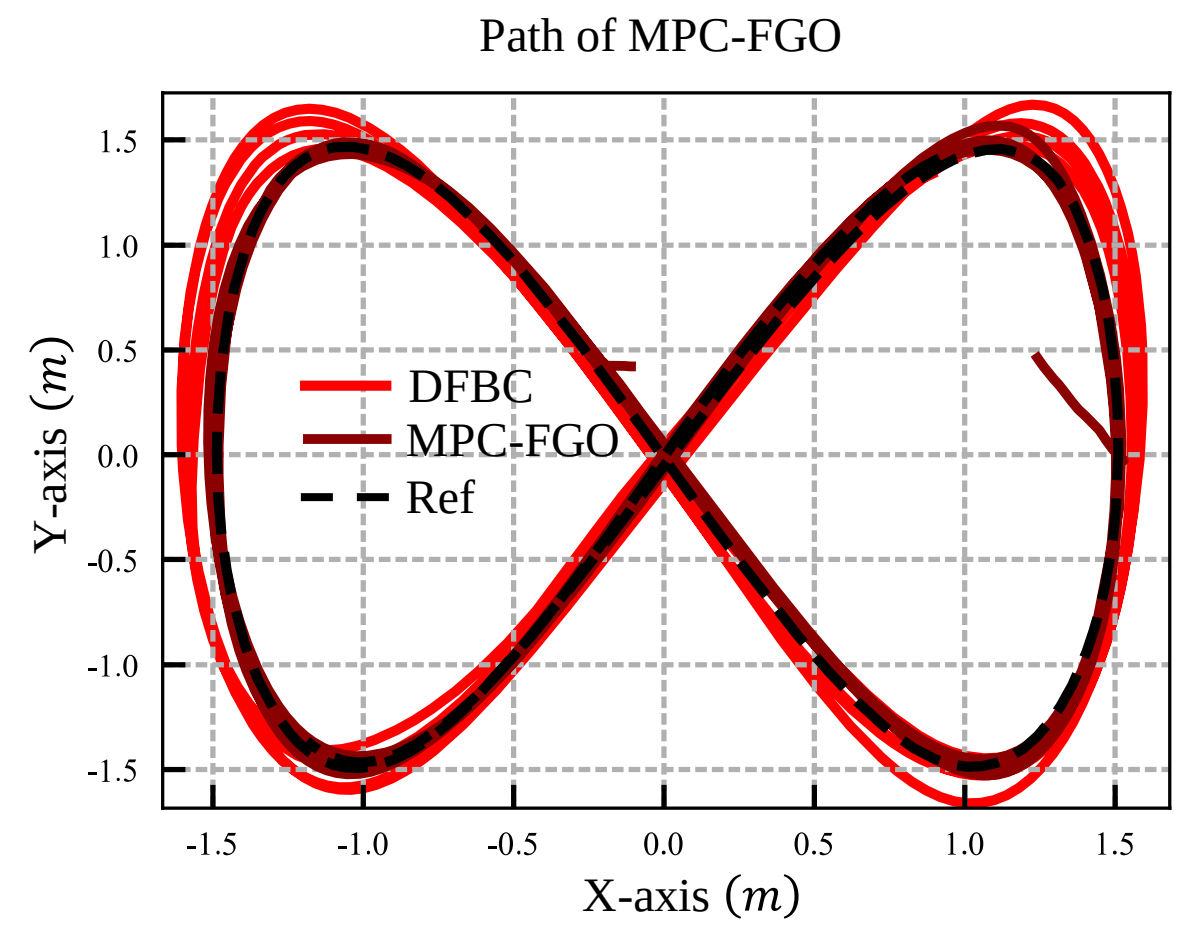

Fig 12. Path tracking based on MPC-FGO with precise positioning. The label Ref represents the reference path.

2) *JPCM versus the traditional model with highly uncertain measurements*

The ground truth from the motion capture system is used as the Ref. Gaussian noise, with a sigma of 0.25m for each positional component, is added to the ground truth to simulate GNSS position measurements. As shown in Fig 13 (a), the JPCM shows a lower tracking error than the traditional model. Compared with the traditional model, the proposed JPCM is significantly smoother because the JCPM naturally inherits uncertainty from measurements. However, the traditional method has significant jitters because the traditional model always absolutely trusts the positioning solution from the FGO-based positioning module [12]. However, in experimental scenarios where the positioning uncertainty is relatively high, more aggressive control variables will be introduced. It can be verified in Fig 13 (b), which shows that the thrust of the traditional method will have a larger range of thrust than the proposed method.

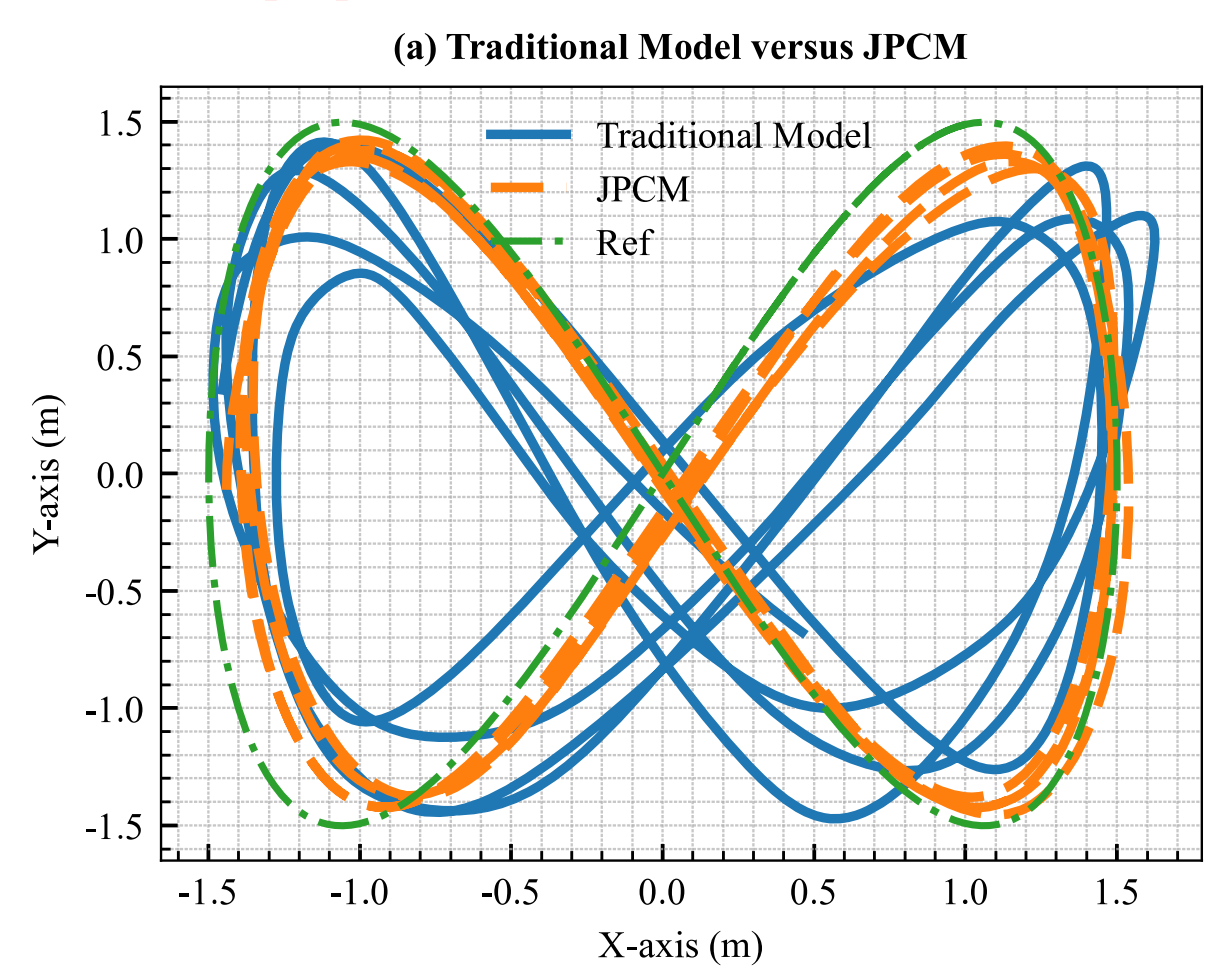

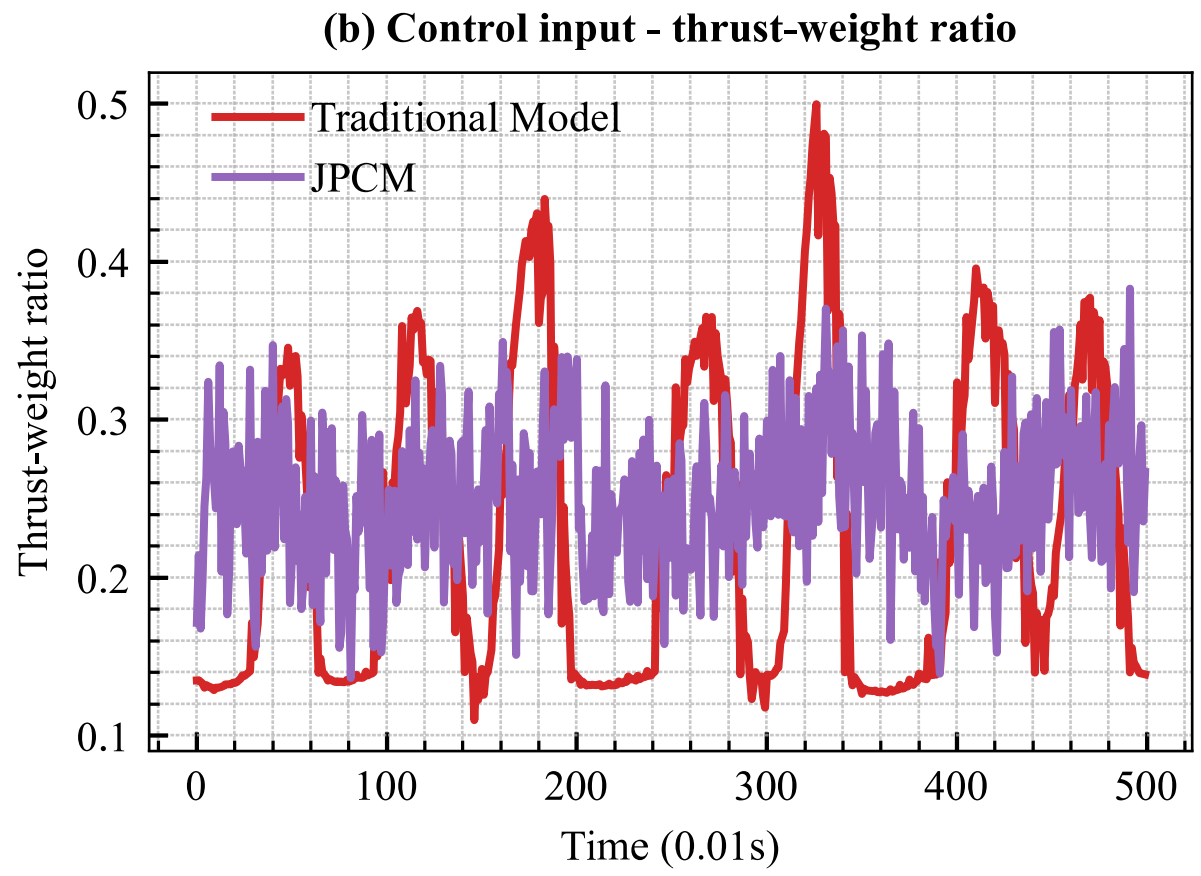

Fig 13. The traditional model versus the proposed JPCM. (a) The path comparison between the traditional model and proposed JPCM. (b) The thrust-weight ratio comparison.

#### *3) Energy efficiency*

In the simulation, as illustrated in Fig 9, the amplitude of the control input is significantly decreased, which means less thrust is needed. In addition, the proposed model can reduce the vibration of the UAV. Therefore, we infer that the proposed model can reduce the energy cost of the whole system. To verify this point, we have collected power data from the controller and have compared the energy cost between the traditional and proposed models. The average power consumption is reduced from 386 W in the traditional model to 364 W in the proposed model. Therefore, studies show that integrating state estimation with control can yield energy savings.

The implications for UAV operational efficiency are substantial. Reduced energy cost means longer flight times, extended operational ranges, and the ability to carry additional payloads without compromising performance. This makes UAVs more efficient and versatile for various applications, from delivery services to environmental monitoring.

### *b) Computational complexity comparison*

The Levenberg–Marquardt algorithm (LMA) [46] is used to solve the factor graph optimization problem of the proposed JPCM or positioning. During optimization, the LMA update is as follows:

$$\begin{gathered}\left(\mathbf{J}^T\mathbf{W}\mathbf{J}+\lambda\mathrm{diag}(\mathbf{J}^T\mathbf{W}\mathbf{J})\right)\delta\boldsymbol{\chi}=\mathbf{J}^T\mathbf{W}\mathbf{r}\\ \text{with } \mathbf{r}=[\mathbf{r}_1^T \quad \dots \quad \mathbf{r}_m^T]^T,\ \delta\boldsymbol{\chi}=\\ \left\{\delta\mathbf{x}_{-M+1:N}^{\mathrm{T}},\delta\mathbf{u}_{0:N-1}^{\mathrm{T}}\right\}^{\mathrm{T}}\in\mathbb{R}^{M+N},\ \mathbf{J}=\frac{\partial\mathbf{r}}{\partial\delta\boldsymbol{\chi}}\end{gathered} \tag{23}$$

where $\mathbf{r}$ is the stacked residuals vector, $\lambda$ is the damping coefficient, $\boldsymbol{\chi}$ is the unknown variable, $\mathbf{J}$ denotes the Jacobian matrix of $\mathbf{r}$ relative to $\delta\boldsymbol{\chi}$, and $\mathbf{W}$ is the Fisher information matrix.

We assume that the initialization of variables is close enough to the global minimum to allow convergence, and the number of iterations is typically small. According to (23), the overall complexity of the LMA method is determined by the number of iterations required for convergence and the complexity of each iteration $\left(\mathbf{J}^T\mathbf{W}\mathbf{J}+\lambda\mathrm{diag}(\mathbf{J}^T\mathbf{W}\mathbf{J})\right)^{-1}\mathbf{J}^T\mathbf{W}\mathbf{r}$.

First, let's examine the complexity of the traditional model. In the traditional model, where the positioning and control problems are solved in turn, the positioning Hessian matrices $\mathbf{H}_p$ and control Hessian matrices $\mathbf{H}_c$ are as follows:

$$\begin{aligned}\mathbf{H}_p&=\mathbf{J}_{\chi_p}\mathbf{W}_p\mathbf{J}_{\chi_p}\\ \mathbf{H}_c&=\mathbf{J}_{\chi_c}\mathbf{W}_c\mathbf{J}_{\chi_c}\end{aligned} \tag{24}$$

To compare the complexity of JCPM and traditional models, we rewrite the Jacobian matrix $\mathbf{J}$ and Fisher information matrix $\mathbf{W}$ as follows:

$$\mathbf{J}=\left[\begin{array}{c|c}\mathbf{J}_{\chi_p} & \mathbf{0}\\ \hline \mathbf{0}\ \ \mathbf{J}_{\delta\mathbf{x}_0}\ \ \mathbf{0} & \mathbf{J}_{\chi_c}\end{array}\right],\ \mathbf{W}=\left[\begin{array}{c|c}\mathbf{W}_p & \begin{matrix}\mathbf{0}\\ \mathbf{W}_0^T\ \ \mathbf{0}\end{matrix}\\ \hline \begin{matrix}\mathbf{0}\ \ \mathbf{W}_0\\ \mathbf{0}\end{matrix} & \mathbf{W}_c\end{array}\right] \tag{25}$$

where $\boldsymbol{\chi}=\{\chi_p,\chi_c\}$, $\chi_p=\{\delta\mathbf{x}_{-M-1:0}\}$, and $\chi_c=\{\delta\mathbf{x}_{1:N},\delta\mathbf{u}_{1:N-1}\}$. $\mathbf{J}_{\delta\mathbf{x}_0}\in\mathbb{R}^{9\times 9}$ denotes the Jacobian matrix of dynamic factor residual to $\delta\mathbf{x}_0$. $\mathbf{W}_0\in\mathbb{R}^{14\times 9}$ denotes the covariance matrix of $\mathbf{x}_0$ to $(\mathbf{x}_1^{\mathrm{T}},\mathbf{u}_1^{\mathrm{T}})^{\mathrm{T}}$.

The matrices $\mathbf{J}$ and $\mathbf{W}$ can be decomposed into a sparse matrix ($\mathbf{J}_1$ and $\mathbf{W}_1$) and another ultra-sparse matrix ($\mathbf{J}_2$ and $\mathbf{W}_2$), respectively, as shown below:

$$\begin{gathered}\mathbf{J}=\mathbf{J}_1+\mathbf{J}_2,\mathbf{J}_1=\begin{bmatrix}\mathbf{J}_{\chi_p} & \mathbf{0}\\ \mathbf{0} & \mathbf{J}_{\chi_c}\end{bmatrix},\mathbf{J}_2=\left[\begin{array}{c|c}\mathbf{0} & \mathbf{0}\\ \hline \mathbf{0}\ \ \mathbf{J}_{\delta\mathbf{x}_0}\ \ \mathbf{0} & \mathbf{0}\end{array}\right]\\ \mathbf{W}=\mathbf{W}_1+\mathbf{W}_2,\mathbf{W}_1=\begin{bmatrix}\mathbf{W}_p & \mathbf{0}\\ \mathbf{0} & \mathbf{W}_c\end{bmatrix},\mathbf{W}_2=\left[\begin{array}{c|c}\mathbf{0} & \begin{matrix}\mathbf{0}\\ \mathbf{W}_0^T\ \ \mathbf{0}\end{matrix}\\ \hline \begin{matrix}\mathbf{0}\ \ \mathbf{W}_0\\ \mathbf{0}\end{matrix} & \mathbf{0}\end{array}\right]\end{gathered} \tag{26}$$

The dimensions of $\mathbf{J}_{\delta\mathbf{x}_0}$ and $\mathbf{W}_0$ are much smaller than $\mathbf{J}_2$ and $\mathbf{W}_2$. Based on (26) and block matrix multiplication, the Hessian matrix of JPCM $\mathbf{H}_m$ is approximately equal to,

$$\mathbf{H}_m=\mathbf{J}^T\mathbf{W}\mathbf{J}\approx\mathbf{J}_1^T\mathbf{W}_1\mathbf{J}_1=\begin{bmatrix}\mathbf{H}_p & \mathbf{0}\\ \mathbf{0} & \mathbf{H}_c\end{bmatrix} \tag{27}$$

Therefore, the complexity of inverting both $\mathbf{H}_p$ and $\mathbf{H}_c$ is approximately equivalent to the Hessian matrix $\mathbf{H}_m$ of the proposed JPCM, and even $\mathbf{H}_p$ and $\mathbf{H}_c$ are sparse matrix [11, 47]. In addition, each iteration can be accelerated by leveraging efficient factorization and intelligent variable reordering, which minimize fill-in and maintain the sparsity of the Hessian matrix [11, 47]. Therefore, the computational complexity of the proposal JPCM is comparable to that of the traditional model.

*In experiments, the proposed JPCM achieves a control frequency of 100 Hz. The average time cost is 5.4 milliseconds of the JPCM, and 99.8% of the optimization problem solving is less than 10ms. The time cost of the traditional model is close to the proposed JPCM.* The relationship curve between time cost and initial position tracking error is shown in Fig 14, showing that in the trajectory tracking experiment, time cost is divided into two distributions with different average values, with a segmentation point of 7cm. Additionally, the time cost of JPCM does not increase linearly with position tracking errors, ensuring real-time performance.

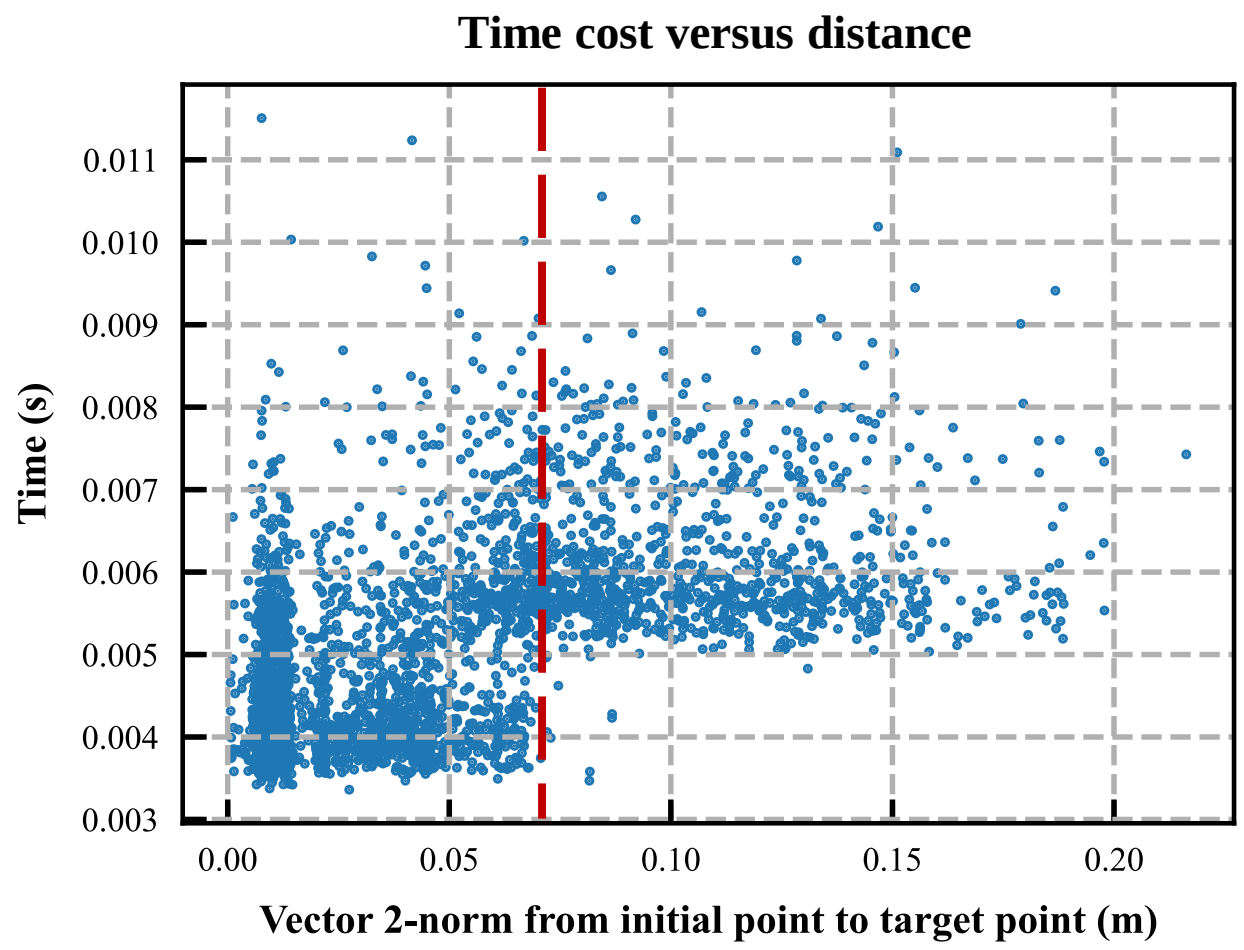

Fig. 14. The time cost of JPCM for different distances between the initial position and the target position.

*c) Optimization techniques analysis*

Initialization, damping coefficients, and termination conditions influence the convergence together. We will discuss these three points in the following.

1) Initialization

Accurate parameter initialization, where the optimal solution is close to the initial value, enhances convergence rate. At the beginning of the optimization process, the latest state is propagated using the unbiased IMU measurement values,

$$\begin{aligned} \mathbf{p}_b^w(t+\Delta t) &= \mathbf{p}_b^w(t) + \mathbf{v}_b^w(t)\Delta t \\ \mathbf{v}_b^w(t+\Delta t) &= \mathbf{v}_b^w(t) + \mathbf{R}_b^w(t)\big(\mathbf{a}(t) - \mathbf{b}_a(t)\big)\Delta t - \mathbf{g}\Delta t \\ \mathbf{R}_b^w(t+\Delta t) &= \mathbf{R}_b^w(t)\mathrm{Exp}\Big(\big(\boldsymbol{\omega}(t) - \mathbf{b}_\omega(t)\big)\Delta t\Big) \end{aligned} \tag{28}$$

where $\mathbf{a}(t)$ and $\boldsymbol{\omega}(t)$ are acceleration and angular velocity from the IMU sensor, $\mathbf{b}_a(t)$ and $\mathbf{b}_\omega(t)$ are the IMU bias that is obtained in the last optimization. In addition, the reference trajectory is regarded as the initial value of control inputs.

2) Damping coefficient

When $\lambda$ is high, the algorithm behaves more like gradient descent, which is more stable but slower. When $\lambda$ is low, the algorithm behaves more like Gauss-Newton, which is faster but can be less stable. We determine whether to adjust the damping coefficient by monitoring error indicators such as residual sum of squares $\psi(\boldsymbol{\chi})$ in [46],

$$\psi(\boldsymbol{\chi}) = \mathbf{r}^T\mathbf{W}\mathbf{r} \tag{29}$$

A metric $\rho(\delta\boldsymbol{\chi})$ [46] is a measure to evaluate convergence performance,

$$\rho_k(\delta\boldsymbol{\chi}) = \frac{\psi(\boldsymbol{\chi}) - \psi(\boldsymbol{\chi}+\delta\boldsymbol{\chi})}{|\delta\boldsymbol{\chi}^T(\lambda_k \mathrm{diag}(\mathbf{J}^T\mathbf{W}\mathbf{J})\delta\boldsymbol{\chi} + \mathbf{J}^T\mathbf{W}\mathbf{r})|} \tag{30}$$

Then, the parameters update rules are as follows:

$$\begin{cases} \boldsymbol{\chi}+\delta\boldsymbol{\chi} \to \boldsymbol{\chi} \; and \; \dfrac{\lambda}{\lambda_f} \to \lambda \; if \; \rho_k(\delta\boldsymbol{\chi}) > \tau_\lambda \\ \lambda \cdot \lambda_f \to \lambda \; otherwise \end{cases} \tag{31}$$

where $\lambda_f$ is a scale factor, $\tau_\lambda$ is a constant threshold. In our experiments, we use the parameters $\lambda_f = 2$, $\tau_\lambda = 0.01$.

3) Termination conditions

Stop optimizing when any of the following termination conditions are met. The maximum number $k$ of iterations is reached ($k = K$); the $k$-th absolute descent error less than or equal to a certain threshold $\sqrt{\psi_k(\boldsymbol{\chi})} \le \tau_a$; and the $k$-th relative descent error less than or equal to a certain threshold $\left|\sqrt{\psi_k(\boldsymbol{\chi})} - \sqrt{\psi_{k+1}(\boldsymbol{\chi})}\right| \le \tau_r (k \ge 1)$. In our experiments, we use the parameters $K = 10$, $\tau_a = 100$, $\tau_r = 0.01$.

## V. Discussion

When substantial errors occur in the initial state, they can induce significant fluctuations in the predicted trajectory, as evidenced by the conventional MPC in Fig 12 from the simulation experiment. This discrepancy arises because the motion control solution fails to align with the actual position of the UAV, potentially leading to unstable control, and any attitude jitter can pose a significant safety risk. However, a control mechanism that incorporates the initial positioning probability may mitigate the risks due to positioning uncertainty. As shown in Fig 12, the predicted trajectory of JPCM is more in line with the true state of UAV.

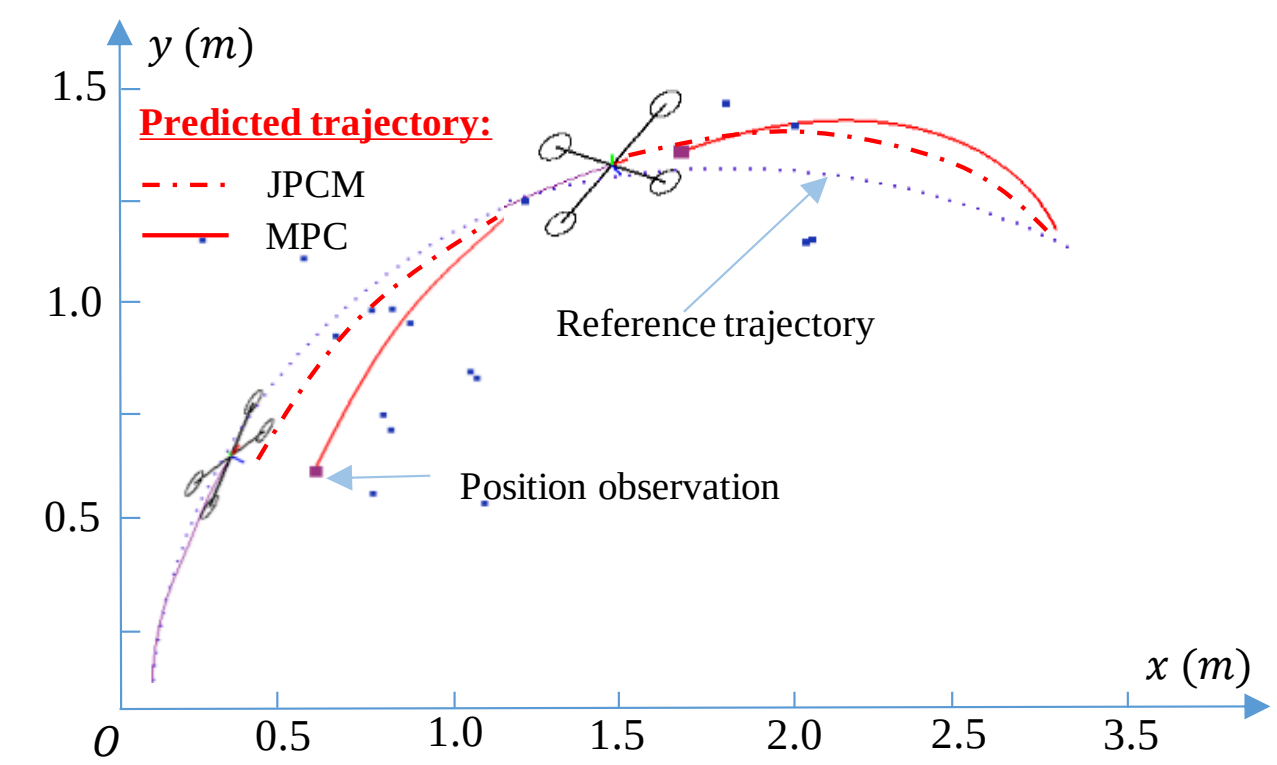

Fig. 12. Predicted trajectory by MPC and JPCM.

When the positioning accuracy is high, JPCM is equivalent to the MPC problem. However, in real-world applications, such as autonomous vehicles navigating city streets and cooperative robots performing tasks in indoor environments, high-accuracy positioning may not be feasible, thus bringing significant challenges to implementing these advanced technologies. In contrast, JPCM is equivalent to relaxing the fixed-value constraints on the initial state in the MPC problem into a probabilistic constraint. This relaxation allows for greater flexibility in the control, potentially improving the UAV's ability to handle unexpected changes in its environment or initial state.

Moreover, the recovery process demonstrates that the proposed JPCM necessitates a longer duration to revert to the pre-set path compared to the conventional MPC. This can be attributed to the simultaneous optimization of positioning and control. The constraints imposed by the planned trajectory exert a pull on the estimated state, drawing it closer to the trajectory, which, in turn, diminishes the energy cost. The proposed JPCM exhibits resilience to faults triggered by irregular positioning or sudden, large-scale movements. There exists a trade-off between the response rate and fault resilience. Intuitively, the higher the state uncertainty, the lower the confidence of the control in state estimation. But sometimes it is necessary to increase the response rate, and we can also appropriately decrease the covariance matrix of the state when the deviation is too large. Therefore, we can choose between response rate and resilience by changing the weighting matrices.

*Potential for real-world application:* (1) The proposed model can be applied to different types of UAVs or larger swarms of UAVs. The dynamic model factor needs to be adjusted to match the corresponding dynamic model of the specific robot. (2) UAV applications in highly urbanized areas are challenged due to dynamic environments and objects, which can introduce significant uncertainties in both positioning and control. These dynamic objects not only introduce association errors but also increase the probability of collisions. Furthermore, the presence of numerous obstacles requires robust and real-time path planning algorithms to ensure safe navigation and obstacle avoidance. Additionally, signal interference due to dense infrastructure and electronic devices can affect the reliability of GNSS and communication signals.

## VI. Conclusion and Future Work

To establish a tightly integrated model with positioning and control as an FGO problem, we propose a tight JPCM. In addition, we provide a unified design framework for the unified factor graph, formulating a series of factors pertinent to positioning, dynamics constraints, and trajectory tracking. These factors are instrumental in ensuring precise and efficient control over the pre-planned trajectory. In simulations, the proposed method is tested under disturbances such as abrupt movements and aerodynamic drag, demonstrating better resilience compared to traditional MPC. The simulation results show that the proposed method is convergent with smaller errors than the conventional MPC. Furthermore, trajectory tracking experiments using a quadrotor also are performed to evaluate feasibility and effectiveness, demonstrating real-time and robust tracking performance. The complexity and optimization techniques are also discussed. Future work should focus on developing safety-quantifiable monitoring to ensure the safety of the proposed JCPM. In addition, integrating more sensors in the JPCM and testing in urbanized areas, will be essential.

## VII. Acknowledgment

This paper is funded by the MEITUAN ACADEMY OF ROBOTICS SHENZHEN under the project “Vision Aided GNSS-RTK Positioning for UAV System in Urban Canyons (ZGHQ)”. This paper is also funded by the PolyU Research Institute for Advanced Manufacturing (RIAM) under the project “Unmanned Aerial Vehicle Aided High Accuracy Addictive Manufacturing for Carbon Fiber Reinforced Thermoplastic Composites Material (CD8S)”.

## VIII. Appendix

Jacobian matrix of dynamic factors

The Jacobian matrix of error $\mathbf{e}^D$ to state $\mathbf{x}_i$ is as follows:

$$J_{\mathbf{x}_i}^{\mathbf{e}^D} = \begin{bmatrix} -\mathbf{R}_w^{b_i} m_b & \left(\mathbf{R}_w^{b_i}\mathbf{p}_p\right)^{\times} & -\mathbf{R}_w^{b_i} m_b \Delta t & \mathbf{0} \\ \mathbf{0} & -\mathbf{R}_w^{b_{i+1}}\mathbf{R}_{b_i}^w & \mathbf{0} & -\mathbf{I}_{3\times3}\Delta t \\ \mathbf{0} & \frac{\partial \bar{\mathbf{e}}_v}{\partial \mathbf{R}_{b_i}^w} & \frac{\partial \bar{\mathbf{e}}_v}{\partial \mathbf{v}_{b_i}^w} & \mathbf{0} \\ \mathbf{0} & \mathbf{0} & \mathbf{0} & \frac{\partial \bar{\mathbf{e}}_\omega}{\partial \boldsymbol{\omega}_{b_i}} \end{bmatrix} \tag{32}$$

where $\mathbf{p}_p$, $\mathbf{p}_v$, $\frac{\partial \bar{\mathbf{e}}_v}{\partial \mathbf{R}_{b_i}^w}$, $\frac{\partial \bar{\mathbf{e}}_v}{\partial \mathbf{v}_{b_i}^w}$, and $\frac{\partial \bar{\mathbf{e}}_\omega}{\partial \boldsymbol{\omega}_{b_i}}$ are as follows:

$$\mathbf{p}_p = m_b\left(\mathbf{p}_{b_{i+1}}^w - \mathbf{v}_{b_i}^w \Delta t + 0.5\mathbf{R}_G^w \mathbf{e}_3 g \Delta t^2 - \mathbf{p}_{b_i}^w\right) \tag{33a}$$

$$\mathbf{p}_v = m_b\left(\mathbf{v}_{b_{i+1}}^w - \mathbf{v}_{b_i}^w + \mathbf{R}_G^w \mathbf{e}_3 g \Delta t\right) \tag{33b}$$

$$\frac{\partial \bar{\mathbf{e}}_v}{\partial \mathbf{R}_{b_i}^w} = \left(\mathbf{R}_w^{b_i}\mathbf{p}_v\right)^{\times} - \mathbf{D}\left(\mathbf{R}_w^{b_i}\mathbf{v}_{b_i}^w\right)^{\times}\Delta t \tag{33c}$$

$$\frac{\partial \bar{\mathbf{e}}_v}{\partial \mathbf{v}_{b_i}^w} = -\mathbf{R}_w^{b_i} m_b - \mathbf{D}\mathbf{R}_w^{b_i}\Delta t \tag{33d}$$

$$\frac{\partial \bar{\mathbf{e}}_\omega}{\partial \boldsymbol{\omega}_{b_i}} = \begin{bmatrix} -I_b^1 & \left(I_b^3 - I_b^2\right)\omega_3 & \left(I_b^3 - I_b^2\right)\omega_2 \\ \left(I_b^1 - I_b^3\right)\omega_3 & -I_b^2 & \left(I_b^1 - I_b^3\right)\omega_1 \\ \left(I_b^2 - I_b^1\right)\omega_2 & \left(I_b^2 - I_b^1\right)\omega_1 & -I_b^3 \end{bmatrix}\Delta t \tag{33e}$$

The Jacobian matrix of error $\mathbf{e}^D$ to state $\mathbf{x}_{i+1}$ is as follows:

$$J_{\mathbf{x}_{i+1}}^{\mathbf{e}^D} = \begin{bmatrix} \mathbf{R}_w^{b_i} m_b & \mathbf{0} & \mathbf{0} & \mathbf{0} \\ \mathbf{0} & \mathbf{I}_{3\times3} & \mathbf{0} & \mathbf{0} \\ \mathbf{0} & \mathbf{0} & \mathbf{R}_w^{b_i} m_b & \mathbf{0} \\ \mathbf{0} & \mathbf{0} & \mathbf{0} & \mathbf{I}_b \end{bmatrix} \tag{34}$$

The Jacobian matrix of error $\bar{\boldsymbol{e}}_v$ to control input $\mathbf{u}$ is as follows:

$$J_{\mathbf{u}}^{\bar{\mathbf{e}}_v} = J_{\mathbf{T}}^{\bar{\mathbf{e}}_v} J_{\mathbf{u}}^{\mathbf{T}} = -\Delta t \begin{bmatrix} \mathbf{0}_4^T \\ \mathbf{0}_4^T \\ 2c_t\mathbf{u}^T \end{bmatrix} \tag{35}$$

The Jacobian matrix of error $\bar{\boldsymbol{e}}_{\boldsymbol{\omega}}$ to control input $\mathbf{u}$ is as follows:

$$\begin{aligned} J_{\mathbf{M}}^{\bar{\mathbf{e}}_\omega} &= J_{\mathbf{M}}^{\bar{\mathbf{e}}_\omega} J_{\mathbf{u}}^{\mathbf{M}} \\ &= -2c_t\Delta t \begin{Bmatrix} \left[\mathbf{p}_{r_1}^b \times \mathbf{e}_3 \quad \cdots \quad \mathbf{p}_{r_K}^b \times \mathbf{e}_3\right] \\ +\left[k_m\mathbf{e}_3 \quad \cdots \quad (-1)^K k_m \mathbf{e}_3\right] \end{Bmatrix} \begin{bmatrix} \mathbf{u}^T \\ \mathbf{u}^T \\ \mathbf{u}^T \end{bmatrix} \end{aligned} \tag{36}$$

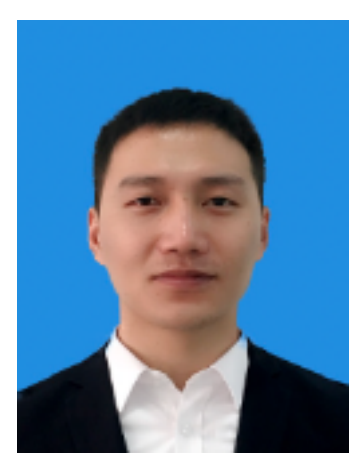

**Peiwen Yang** received the M.S. degree in the School of Information and Electronics, Beijing Institute of Technology, Beijing, China, in 2019. He is currently a Ph.D. student in the Department of Aeronautical and Aviation Engineering, at the Hong Kong Polytechnic University.

His current research interests include aerial vehicle control, computer vision, and robotics.

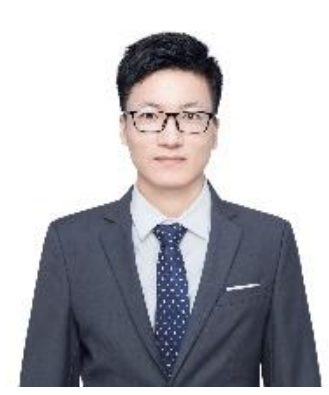

**Weisong Wen** (Member, IEEE) received a BEng degree in Mechanical Engineering from Beijing Information Science and Technology University (BISTU), Beijing, China, in 2015, and an MEng degree in Mechanical Engineering from the China Agricultural University, in 2017. After that, he received a PhD degree in Mechanical Engineering from The Hong Kong Polytechnic University (PolyU), in 2020. He was also a visiting PhD student with the Faculty of Engineering, University of California, Berkeley (UC Berkeley) in 2018. Before joining PolyU as an Assistant Professor in 2023, he was a Research Assistant Professor at AAE of PolyU since 2021. He has published 30 SCI papers and 40 conference papers in the field of GNSS (ION GNSS+) and navigation for Robotic systems (IEEE ICRA, IEEE ITSC), such as autonomous driving vehicles. He won the innovation award from TechConnect 2021, the Best Presentation Award from the Institute of Navigation (ION) in 2020, and the First Prize in Hong Kong Section in Qianhai-Guangdong-Macao Youth Innovation and Entrepreneurship Competition in 2019 based on his research achievements in 3D LiDAR aided GNSS positioning for robotics navigation in urban canyons. The developed 3D LiDAR-aided GNSS positioning method has been reported by top magazines such as Inside GNSS and has attracted industry recognition with remarkable knowledge transfer.

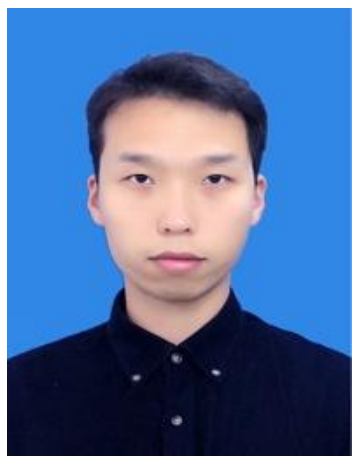

**Shiyu Bai** (Member, IEEE) was born in Xuzhou, Jiangsu, China. He received a Ph.D. degree in navigation, guidance, and control from Nanjing University of Aeronautics and Astronautics, Nanjing, China, in 2022. He is currently a postdoctoral fellow in the Department of Aeronautical and Aviation Engineering at the Hong Kong Polytechnic University. His research interests include inertial navigation, multi-sensor fusion, indoor positioning, and vehicular positioning.

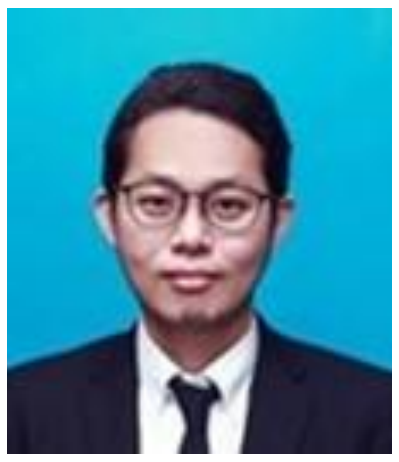

**Li-Ta Hsu** (Senior Member, IEEE) received the B.S. and Ph.D. degrees in aeronautics and astronautics from National Cheng Kung University, Tainan City, Taiwan, in 2007 and 2013, respectively. He is currently an Associate Professor at the Department of Aeronautical and Aviation Engineering, The Hong Kong Polytechnic University. He is also Limin Endowed Young Scholar in Aerospace Navigation. He was a Visiting Researcher with the Faculty of Engineering, University College London, London, U.K., and the Tokyo University of Marine Science and Technology, Tokyo, Japan, in 2012 and 2013, respectively. Dr. Hsu was selected as a Japan Society for the Promotion of Sciences Postdoctoral Fellow with the Institute of Industrial Science, University of Tokyo, Tokyo, Japan, and worked from 2014 to 2016. He is an Associate Fellow with the Royal Institute of Navigation, London, U.K. He is currently a member of ION and a member of the editorial board and reviewer in professional journals related to GNSS. In 2013, he won a Student Paper Award and two Best Presentation Awards from the Institute of Navigation (ION).

His research interests include GNSS positioning in challenging environments and localization for pedestrians, autonomous driving vehicles, and unmanned aerial vehicles.